\documentclass{article}

\usepackage{microtype}
\usepackage{graphicx}
\usepackage{subfigure}
\usepackage{booktabs} 

\usepackage{multirow}
\usepackage{graphicx}
\usepackage{titlesec}
\usepackage{diagbox} %
\usepackage{appendix}
\usepackage{bbm}
\usepackage[table]{xcolor}
\usepackage{threeparttable}
\usepackage{bbding}
\usepackage{makecell}
\usepackage{lscape}
\usepackage{tabularx}

\usepackage{hyperref}

\definecolor{MyRed}{RGB}{204,51,17}
\definecolor{MyBlue}{RGB}{0,119,187}

\usepackage[accepted]{icml2024}

\usepackage{amsmath}
\usepackage{amssymb}
\usepackage{mathtools}
\usepackage{amsthm}

\usepackage[capitalize,noabbrev]{cleveref}

\theoremstyle{plain}

\theoremstyle{definition}

\theoremstyle{remark}

\usepackage[textsize=tiny]{todonotes}

\icmltitlerunning{CLIF: Complementary Leaky Integrate-and-Fire Neuron for Spiking Neural Networks}

\begin{document}

\twocolumn[
\icmltitle{CLIF: Complementary Leaky Integrate-and-Fire Neuron \\ for Spiking Neural Networks}



\icmlsetsymbol{equal}{*}

\begin{icmlauthorlist}
\icmlauthor{Yulong Huang}{equal,hkust}
\icmlauthor{Xiaopeng Lin}{equal,hkust}
\icmlauthor{Hongwei Ren}{hkust}
\icmlauthor{Haotian Fu}{hkust}
\icmlauthor{Yue Zhou}{hkust}
\icmlauthor{Zunchang Liu}{hkust}
 \\

\icmlauthor{Biao Pan}{beihang}
\icmlauthor{Bojun Cheng}{hkust}
\end{icmlauthorlist}

\icmlaffiliation{hkust}{Function Hub, The Hong Kong University of Science
and Technology (Guangzhou), Guangzhou, China}
\icmlaffiliation{beihang}{School of Integrated Circuit Science and Engineering, Beihang University, Beijing, China}

\icmlcorrespondingauthor{Bojun Cheng}{bocheng@hkust-gz.edu.cn}

\icmlkeywords{Spiking Neural Networks, Neuron Model}

\vskip 0.3in
]



\printAffiliationsAndNotice{\icmlEqualContribution} 


\begin{abstract}
Spiking neural networks (SNNs) are promising brain-inspired energy-efficient models. Compared to conventional deep Artificial Neural Networks (ANNs), SNNs exhibit superior efficiency and capability to process temporal information. However, it remains a challenge to train SNNs due to their undifferentiable spiking mechanism. The surrogate gradients method is commonly used to train SNNs, but often comes with an accuracy disadvantage over ANNs counterpart. We link the degraded accuracy to the vanishing of gradient on the temporal dimension through the analytical and experimental study of the training process of Leaky Integrate-and-Fire (LIF) Neuron-based SNNs. Moreover, we propose the Complementary Leaky Integrate-and-Fire (CLIF) Neuron. CLIF creates extra paths to facilitate the backpropagation in computing temporal gradient while keeping binary output. CLIF is hyperparameter-free and features broad applicability. Extensive experiments on a variety of datasets demonstrate CLIF's clear performance advantage over other neuron models. Furthermore, the CLIF's performance even slightly surpasses superior ANNs with identical network structure and training conditions.
The code is available at \href{https://github.com/HuuYuLong/Complementary-LIF}{https://github.com/HuuYuLong/Complementary-LIF}.
\end{abstract}

\section{Introduction}
Spiking Neural Networks (SNNs) \cite{maass1997networks} have captivated the attention of both academic and industrial communities in recent years \cite{tavanaei2019deep, roy2019towards, mehonic2022brain, schuman2022opportunities}. Drawing inspiration from the biological neuron, SNNs adopt the spiking neuron, like the leaky integrate and fire (LIF) model, utilizing spike-based communication for information transmission \cite{teeter2018generalized}. This fundamental characteristic equips SNNs with the capacity to effectively process information across both temporal and spatial dimensions, excelling in areas of low latency and low power consumption \cite{schuman2022opportunities}. Compared to conventional deep Artificial Neural Networks (ANNs), SNNs exhibit superior efficiency and capability to process temporal information, presenting significant implementation potential in edge device applications for real-time applications \cite{roy2019towards, mehonic2022brain, shen2024efficient}.

Despite the advantages of SNNs, the training of SNNs presents a substantial challenge due to the inherently undifferentiable spiking mechanism. Many scholars have intensively explored this problem, three mainstream training methods have been proposed: the bio-inspired training method \cite{kheradpisheh2018stdp}, the ANN-to-SNN conversion method \cite{deng2020optimal} and the surrogate gradient (SG) method \cite{li2021free, wang2022signed, jiang2023unified}. The bio-inspired training method bypasses undifferentiable problems by calculating the gradients with respect to the spike time \cite{zhang2018brain,dong2023unsupervised}. The ANN-to-SNN method utilizes pre-trained ANN models to approximate the ReLU function with the spike neuron model \cite{li2021free, jiang2023unified}. The SG method uses surrogate gradients to approximate the gradients of non-differentiable spike functions during backpropagation \cite{neftci2019surrogate}. This method solves the problem of non-differentiable spike functions, facilitating the direct trainable ability of SNNs \cite{xu2023constructing}. 

\begin{figure*}[ht!]
\label{main picture}
\centering
\includegraphics[
width= 1. \textwidth,
trim=0 50 0 50,clip]{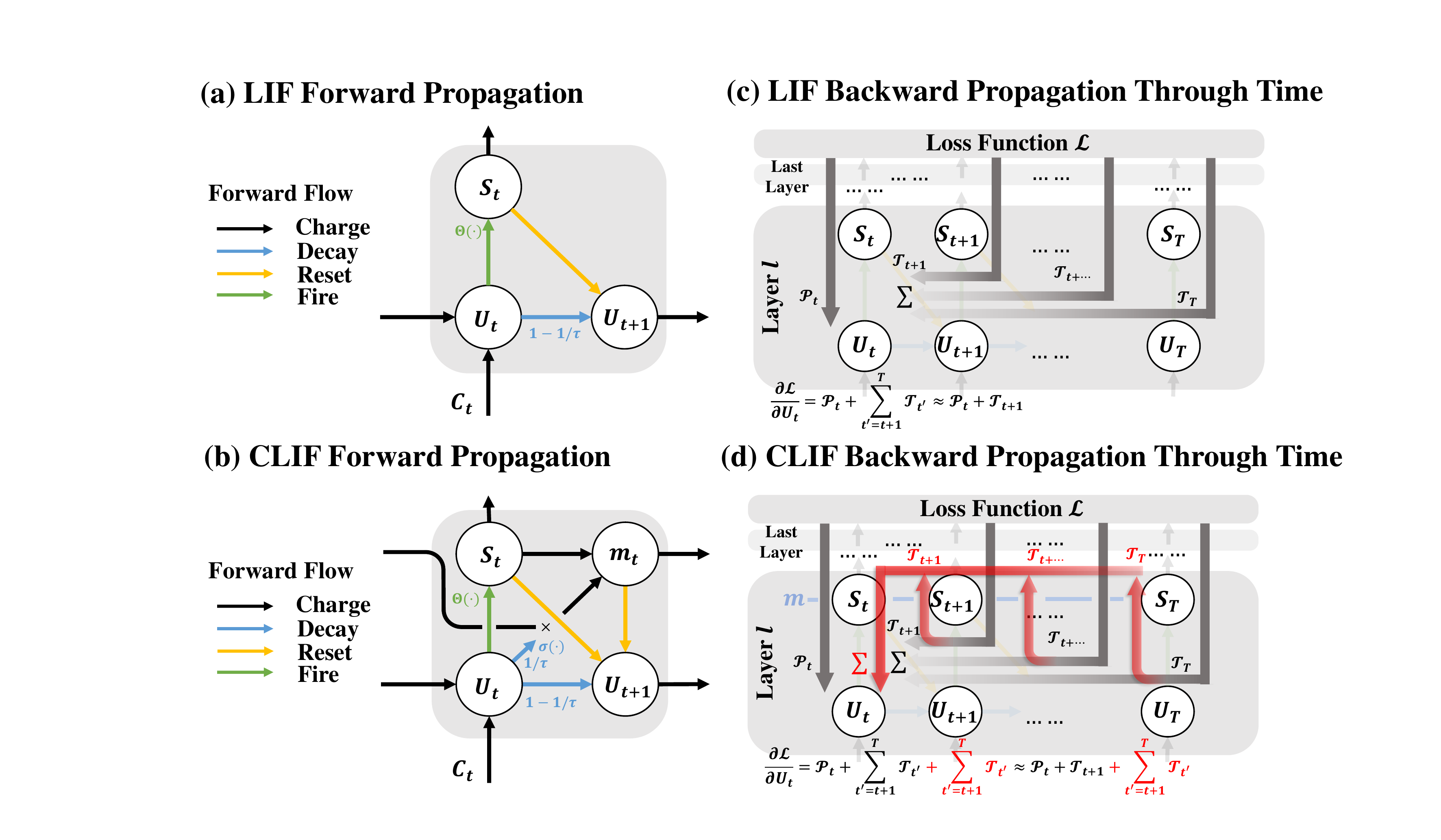}  
\caption{(a) Illustration of the LIF neuron model with forward propagation data flow (b) Illustration of the CLIF neuron model with forward propagation data flow (c) Illustration of the LIF’s gradient error  $\frac{\partial\mathcal{L}}{\partial \boldsymbol{u}^{l}[t]}$ flow during BPTT. Each path is represented by an arrow. Lighter color in the arrow indicates more decay of gradient error. (d) Illustration of the CLIF’s gradient error flow during BPTT. Compared to (c), the additional temporal gradient error is highlighted in red.}
\label{fig:main_pitcture}
\end{figure*}

Each method is attractive in certain aspects but also process certain limitations. SG and ANN-to-SNN methods provide great applicability across various neural network architectures, such as spike-driven MLP \cite{li2022brain}, SRNN \cite{zhang2021skip}, SCNN \cite{fang2021deep} and Transformer backbone \cite{zhou2022spikformer, yao2024spike}. In contrast, bio-inspired training is challenging to be effectively applied to deeper network configurations \cite{kheradpisheh2018stdp}. SG can reach satisfactory performance within limited timestep, whereas ANN-to-SNN requires a large number of timestep and more spikes to achieve comparable accuracy to the network trained by the SG method \cite{deng2020rethinking}. As such, SG-based SNNs are more attractive in edge scenarios where the inference power is critical \cite{roy2019towards}. Nevertheless, the SG method necessitates the use of inaccurate approximations for computing the gradients, leading to imprecise gradient update values and thus diminishing accuracy \cite{wang2023adaptive}.

In this study, we rethink the SNN training process and introduce complementary leaky integrate and fire (CLIF) neuron. The LIF and CLIF neuron model is illustrated in Figure.\ref{fig:main_pitcture}(a) and (b). We introduce a complementary membrane potential ($\boldsymbol{m}[t]$) in CLIF neuron. The complementary potential captures and maintains information related to the decay of the membrane potential. CLIF creates extra paths to facilitate the data flow in temporal gradient computation, as intuitively seen in Figure.\ref{fig:main_pitcture} (c) and (d). Our experiments demonstrate that for SNNs with vanilla LIF neurons, employing a limited number of temporal gradients can yield comparable accuracy to those achieved by using gradients across much more temporal steps. Our theoretical analysis reveals such limitation is linked to the vanishing of certain temporal gradients. Experiments show CLIF can boost the SNN performance significantly in both static images and dynamic event streams. Impressively, even with moderate timestep to keep SNN’s low power advantage, CLIF-based SNN achieves comparable or even superior performance to ANN counterpart with identical network structure and training conditions. Our main contributions are:
\begin{itemize}
  \item {We propose the CLIF neuron model to efficiently and accurately extract temporal gradients. The model has zero hyper-parameters and can interchange with LIF neuron in many mainstream SNNs.}
  \item {We demonstrate that CLIF effectively boosts the SNN performance by simply replacing LIF with CLIF. For different SNNs architectures like spiking VGG and Resnet, up to 2\% accuracy improvement is observed.}  
  \item {We conduct extensive experiments and discover that even with moderate timestep to keep the low power advantage, CLIF-based SNNs achieve comparable performance to ANNs with identical network structure and training conditions.}
\end{itemize}

\section{Related Work}
In SG method, gradients of non-differentiable spike functions are approximated by some surrogate gradients during backpropagation, this method enables SNNs to be trained directly by BPTT \cite{werbos1990backpropagation}. However, the inaccurate approximations for computing the gradients cause imprecise gradient update \cite{wang2023adaptive}, and degradation in accuracy. Moreover, as the BPTT method requires iteration of recursive computation over timestep, the training cost grows substantially over large timestep \cite{wu2018spatio}.

To improve the accuracy of the SG method, many efforts have been made. Some studies have advanced the surrogate functions. ASGL method \cite{wang2023adaptive} introduced an adaptive smoothing gradient to reduce gradient noise. LocalZO \cite{mukhoty2023direct} proposed the zeroth-order method to estimate the gradients for neuron. SML method \cite{deng2023surrogate} introduces ANNs module to reduce the gradient noise accumulation when training. Alternatively, enhanced neuron dynamics could also yield in higher SNNs accuracy. For example, PLIF \cite{fang2021incorporating}, LTMD \cite{wang2022ltmd} and GLIF \cite{yao2022glif} introduced learnability in membrane potential, neuron threshold, and different channels, respectively. Nevertheless, even with those efforts, there is still a performance gap between SNNs and ANNs when implemented with identical network architecture. To enhance the training efficiency, several efficient training methods have been proposed. For instance, e-prop \cite{bellec2020solution} entirely discards the temporal gradients, and only uses the gradients of spatial dimension for training. SLTT \cite{meng2023towards} also discards the gradient of the temporal dimension, but randomly chooses a few gradient paths along the spatial dimension. Surprisingly, even after discarding the gradients in the temporal dimension, these methods still obtain comparable performance to the original BPTT approach. We investigate further such counterintuitive phenomena through experiments and conclude the temporal gradient decays too rapidly over multiple timesteps. Details about this observation are given in methods.

To tackle the rapid temporal gradient decay in SNNs, {\color{black}\cite{lotfi2020long} and \cite{xu2024enhancing} proposed spiking LSTM and spiking ConvLSTM in respectively}. Spiking (Conv)LSTM inherits LSTM’s advantage and avoids rapid temporal gradient decay. However, Spiking (Conv)LSTM comes with a significant number of training parameters compared to LIF within each neuron, complicating the network structuring and increasing training effort. Moreover, Spiking (Conv)LSTM restricts the neuron from the critical operation of decay and reset. {\color{black} \cite{dampfhoffer2022investigating, dampfhoffer2023leveraging} proposed spikeGRU preserves the reset process of spike neuron. The SpikeGRU also inherits the gating mechanism of GRU to avoid fast temporal gradient decay, and still keep the number of training parameters}. \cite{fang2024parallel} increased the parallel connection with trainable parameters between the spiking neurons to learn the long-term dependencies. However, this method also restricts the neuron from reset operation and increases the computation complexity. As such, both methods lose the generosity of SNNs and dilute the high efficiency and low power consumption advantage of SNNs. 

In parallel, several bio-inspired models have been developed, transitioning from biological concepts to neuronal model implementations, with the goal of addressing long-term dependency learning issues. For example, the AHP neuron \cite{rao2022long} inspired by after-hyperpolarizing currents, the TC-LIF model \cite{zhang2024tc} inspired by the Prinsky-Rinzel pyramidal neuron and the ELM model \cite{spieler2023expressive} inspired by the cortical neuron. However, few works demonstrate the potential to apply bio-inspired neuron models on large and complex networks. In summary, the methods to improve the temporal gradients not only add significant training complexity but also cannot be generalized to various network backbones.

\section{Preliminary}
The specific notations used in this paper are described in the Appendix.\ref{apx: Notation in the Paper}.
\subsection{SNN Neuron Model}
In the field of SNNs, the most common neuron model is the Leaky Integrate-and-Fire (LIF) model with iterative expression, as detailed in \cite{wu2018spatio}. At each time step $t$, the neurons in the $l$-th layer integrate the postsynaptic current $\boldsymbol{c}^l[t]$ with their previous membrane potential $\boldsymbol{u}^{l}[t-1]$, the mathematical expression is illustrated in Eq.\eqref{eq:preliminary:ut = ut-1}:
\begin{equation}\label{eq:preliminary:ut = ut-1}
\boldsymbol{u}^{l}[t] = (1 - \frac{1}{\tau}) \boldsymbol{u}^l[t - 1]  + \boldsymbol{c}^l[t], 
\end{equation}
where $\tau$ is the membrane time constant. $\tau > 1$ as the discrete step size is 1. The postsynaptic current $\boldsymbol{c}^l[t] = \boldsymbol{W}^l * \boldsymbol{s}^{l-1}[t]$ is calculated as the product of weights $\boldsymbol{W}^l$ and spikes from the preceding layer $\boldsymbol{s}^{l-1}[t]$, simulating synaptic functionality, with $*$ indicating either a fully connect or convolution's synaptic operation.

Neurons will generate spikes $\boldsymbol{s}^l[t]$ by Heaviside function when membrane potential $\boldsymbol{u}^{l}[t]$ exceeds the threshold $V_{\mathrm{th}}$, as shown in Eq.\eqref{eq:preliminary:s=f(u-vth)}:
\begin{equation}\label{eq:preliminary:s=f(u-vth)}
\boldsymbol{s}^l[t]  = \Theta (\boldsymbol{u}^{l}[t] - V_{\mathrm{th}}) = \begin{cases}
1, & \text{if }  \boldsymbol{u}^l[t] \geq V_{\mathrm{th}} \\
0, & \text{otherwise}
\end{cases} .
\end{equation}
After the spike, the neuron will reset its membrane potential. Two ways are prominent in Eq.\eqref{eq:preliminary:reset}:
\begin{equation}\label{eq:preliminary:reset}
\boldsymbol{u}^{l}[t] =
\begin{cases}
 \boldsymbol{u}^l[t] - V_{\mathrm{th}} \boldsymbol{s}^l[t], & \text{soft reset}  \\
 \boldsymbol{u}^l[t] \odot \left( 1 - \boldsymbol{s}^l[t] \right), & \text{hard reset} 
\end{cases}.
\end{equation}

In this work, we chose the soft reset process because it will keep more temporal information \cite{meng2023towards}.


\subsection{SNN Training with Surrogate Gradient}
In the SG method, gradients are computed through BPTT \cite{wu2018spatio}. This involves considering the temporal dimension, where the gradients at $l$-th layer for all timestep $T$ are calculated as Eq.\eqref{eq:preliminary:dL_dW}:
\begin{equation}\label{eq:preliminary:dL_dW}
\nabla_{\boldsymbol{W}^l}\mathcal{L}=\sum_{t=1}^T \frac{\partial\mathcal{L}}{\partial \boldsymbol{u}^{l}[t]} \frac{\partial \boldsymbol{u}^{l}[t]}{\partial \boldsymbol{W}^{l}} ,l=\mathrm{L},\mathrm{L}-1,\cdots,1  ,
\end{equation}
where $\mathcal{L}$ represents the loss function. We define the $\frac{\partial\mathcal{L}}{\partial \boldsymbol{u}^{l}[t]}$ as the gradient error in this paper, the gradient error can be evaluated recursively:
\begin{equation}\label{eq:preliminary:dL_dU}
\frac{\partial\mathcal{L}}{\partial \boldsymbol{u}^{l}[t]}  = \frac{\partial\mathcal{L}}{\partial \boldsymbol{s}^{l}[t]}  \frac{\partial \boldsymbol{s}^{l}[t]}{\partial \boldsymbol{u}^{l}[t]}
+ \sum_{t^{\prime}=t+1}^{T} \frac{\partial\mathcal{L}}{\partial \boldsymbol{s}^{l}[t^{\prime}]} \frac{\partial \boldsymbol{s}^{l}[t^{\prime}]}{\partial \boldsymbol{u}^{l}[t^{\prime}]}
\prod_{t^{\prime\prime}=1}^{t^{\prime}-t} \boldsymbol{\epsilon}^{l}[t^{\prime}-t^{\prime\prime}] ,
\end{equation}

where the $\boldsymbol{\epsilon}^{l}[t]$ for LIF model can be defined as follows in Eq.\eqref{eq:preliminary:eligibility}: 
\begin{equation}\label{eq:preliminary:eligibility}
\boldsymbol{\epsilon}^{l}[t] \equiv \frac{\partial \boldsymbol{u}^{l}[t+1]}{\partial \boldsymbol{u}^{l}[t]} + \frac{\partial \boldsymbol{u}^{l}[t+1]}{\partial \boldsymbol{s}^{l}[t]} \frac{\partial \boldsymbol{s}^{l}[t]}{\partial \boldsymbol{u}^{l}[t]}.
\end{equation}

In particular, for different layers, we have Eq.\eqref{eq:preliminary:dL_dS}:
\begin{equation}\label{eq:preliminary:dL_dS}
\frac{\partial\mathcal{L}}{\partial \boldsymbol{s}^{l}[t]} = 
\begin{cases} 
\frac{\partial\mathcal{L}}{\partial \boldsymbol{s}^{l}[t]}  & \text{ if } l = \mathrm{L} \\ 
\frac{\partial\mathcal{L}}{\partial \boldsymbol{u}^{l+1}[t]}\frac{\partial{\boldsymbol{u}^{l+1}[t]}}{\partial \boldsymbol{s}^{l}[t]}  & \text{ if } l = \mathrm{L}-1,\cdots, 1
\end{cases},
\end{equation}
where the $\frac{\partial{\boldsymbol{u}^{l+1}[t]}}{\partial \boldsymbol{s}^{l}[t]}=(\boldsymbol{W}^{l+1})^{\top}$. The detailed derivations can be found in the Appendix.\ref{apx: LIF-based BPTT with surrogate gradient}. In addition, the non-differentiable problem is solved by approximating {$\frac{\partial \boldsymbol{s}^{l}[t]}{\partial \boldsymbol{u}^{l}[t]} \approx \mathbb{H}\left(\boldsymbol{u}^{l}[t]\right)$} with the surrogate function $\mathbb{H}(\cdot)$ \cite{neftci2019surrogate}. In this work, we chose the rectangle function \cite{wu2019direct, su2023deep}:
\begin{equation}\label{eq:preliminary:rectangle}
\frac{\partial \boldsymbol{s}^{l}[t]}{\partial \boldsymbol{u}^{l}[t]} \approx \mathbb{H}\left(\boldsymbol{u}^{l}[t]\right) = \frac{1}{\alpha} \mathbbm{1} \left( \left| \boldsymbol{u}^{l}[t] - V_{\mathrm{th}} \right| < \frac{\alpha}{2} \right),
\end{equation}
where $\mathbbm{1}(\cdot)$ served as the indicator function. Following \cite{meng2023towards}, the hyperparameter $\alpha$ is set to $V_{\mathrm{th}}$. In this case, Eq.\eqref{eq:preliminary:eligibility} can be rewritten as: 
\begin{equation}\label{eq:preliminary:epsilon_after}
\boldsymbol{\epsilon}^{l}[t] = \gamma \left( 1 - V_{\mathrm{th}} \mathbb{H}\left(\boldsymbol{u}^{l}[t]\right) \right)
\end{equation}
where $\gamma \triangleq 1 - \frac{1}{\tau}$, resulting $\gamma \in (0, 1)$.

\section{Method}

\subsection{Limitation of LIF-based SNN with SG}\label{section: Limitation of LIF-based SNN with SG}
We investigate the limitation of SG training method with LIF-based SNN through both experimental and theoretical analysis.

\textbf{Experimental Observation:} 
To investigate the relationship between temporal parameters and training performance, we conduct two experiments on LIF-based SNNs using the BPTT method. Our experimental setup utilizes a simple convolutional 5-layer SNN model, suitable to conduct analysis with multiple runs with various parameters.

In the first experiment, we introduce variable $k \in [1, T]$. When calculating the error in the backpropagation, the gradients from further timestep beyond k are discarded ($[k+1, T]$). Figure.\ref{fig:appendx:ob_experiment_setup} in Appendix \ref{apx: The detail of Experimental Observation}. highlight the backpropagation example with $k=2$.

\begin{figure}[ht!]
\centering
\subfigure[Varying $\tau$ and $k$]{\includegraphics[width=.22 \textwidth, 
]{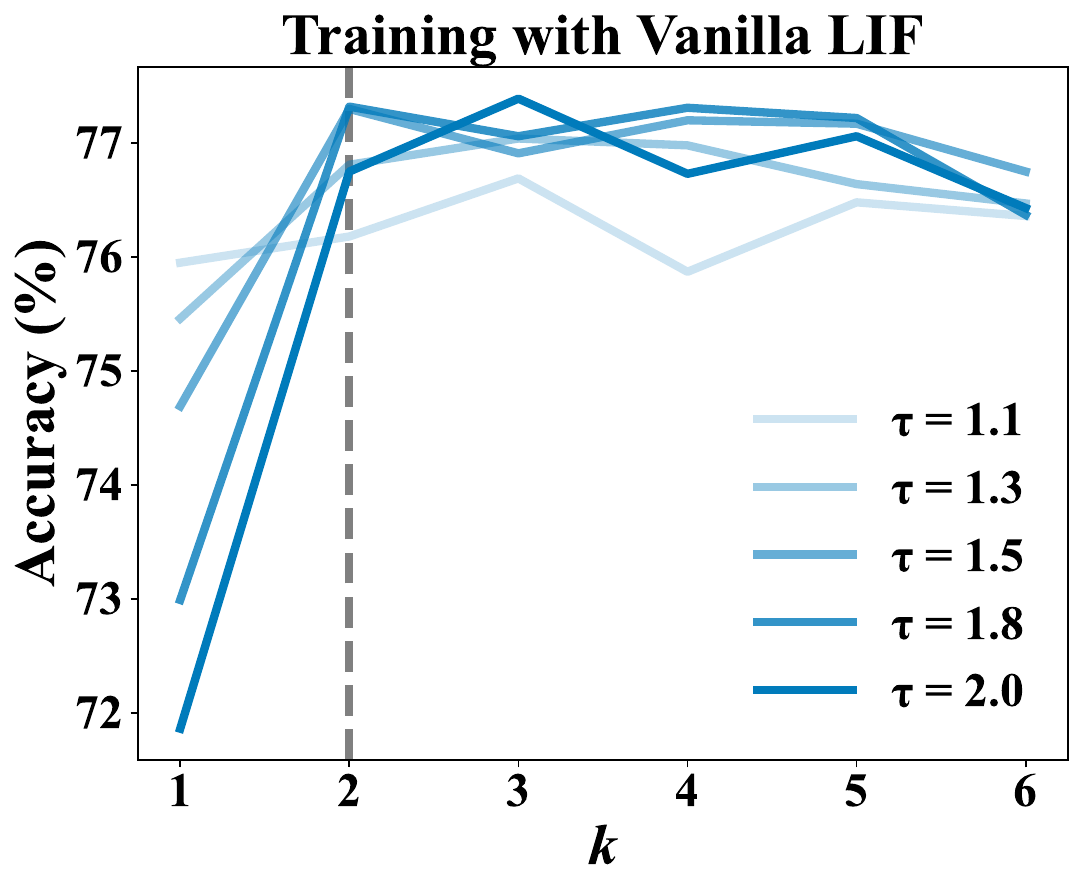}} 
\subfigure[Comparing timestep] {\includegraphics[width=.22 \textwidth, 
]{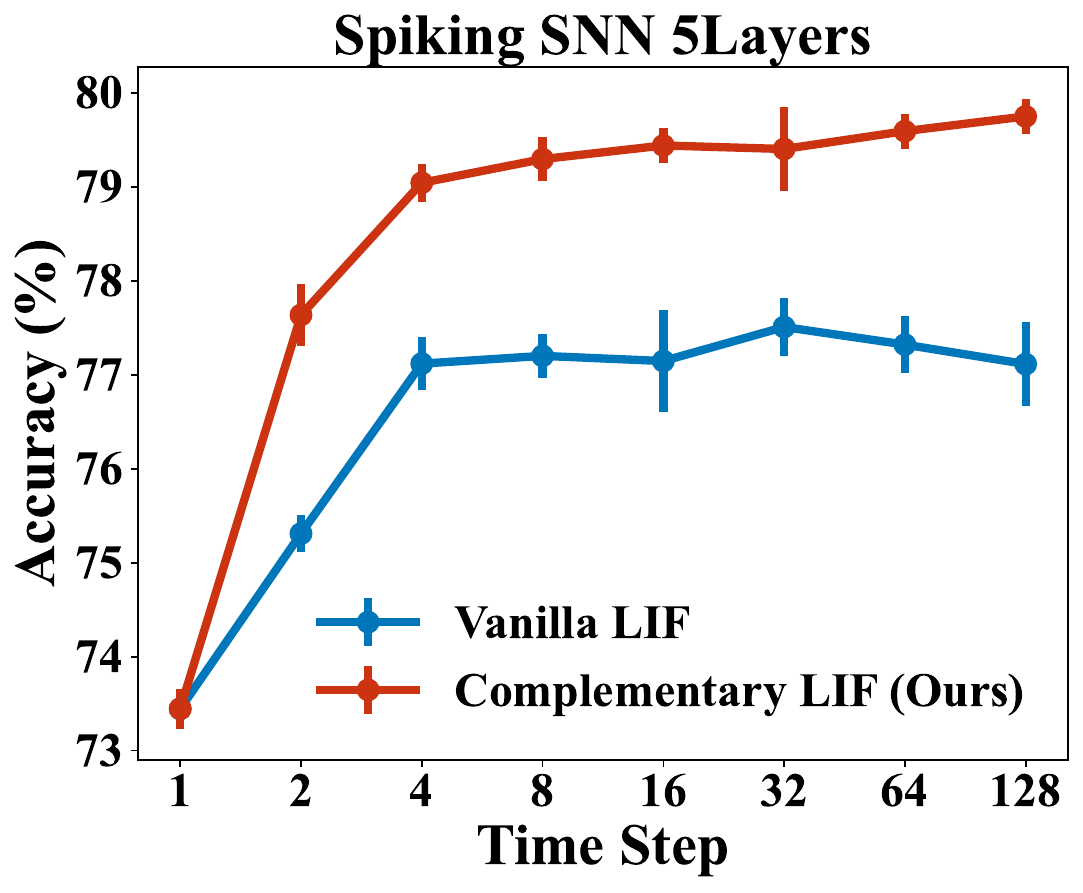}} 
\caption{The performance of LIF based a 5-layer SNN for CIFAR10 dataset: (a) The accuracy is influenced by time constant ($\tau$) and BPTT timestep ($k$) (detaching all gradients from $k+1$ to T during training). We set the timestep to 6. (b) Accuracy over increasing timestep for both vanilla LIF and our proposed CLIF.}
\label{fig:obervervation_tau_vs_k}
\end{figure}

Figure.\ref{fig:obervervation_tau_vs_k}(a) demonstrates how the network accuracy is influenced by time constant ($\tau$) and BPTT timestep ($k$). It appears that the gradient from further timestep could not contribute to the backpropagation training process, as increasing $k$ above 2 does not substantially enhance the accuracy. We repeat this experiment on a few different datasets with different network backbones, all leading to the same conclusion. Experiment results from other datasets are given in Appendix \ref{apx: The detail of Experimental Observation}. 

Figure.\ref{fig:obervervation_tau_vs_k}(b) plots the classification accuracy over increasing timestep for both vanilla LIF and our proposed CLIF. {\color{black}The average and standard error are calculated from the results using 4 different random seeds}. The accuracy of the vanilla LIF peaks at $T=32$ and then declines as the number of timesteps increases. This indicates the temporal gradient from LIF over larger timestep cannot be properly processed. In contrast, the CLIF model demonstrates a sustained improvement of performance over increasing timestep, showcasing CLIF’s effectiveness in learning over longer timestep.

\textbf{Theoretical Analysis:}
Figure.\ref{fig:obervervation_tau_vs_k} reveals LIF’s limitation of exploiting temporal information over a long period. This phenomenon is further investigated analytically. We separate the gradients into spatial $\mathcal{P}^l[t]$ component and temporal component $\sum \mathcal{T}^l[t]$, as shown in \begin{equation}\label{eq:observe:temporal gradients}
\frac{\partial\mathcal{L}}{\partial \boldsymbol{u}^{l}[t]}  
= \mathcal{P}^l[t]
+ \sum_{t^{\prime}=t+1}^{T} \mathcal{T}^l[t, t^{\prime}],
\end{equation}
where $\mathcal{P}^l[t]$ and $\mathcal{T}^l[t, t^\prime]$ can be further expanded as:
\begin{gather}
\label{eq:observe:spatial gradients errors}
\mathcal{P}^l[t] = \frac{\partial\mathcal{L}}{\partial \boldsymbol{s}^{l}[t]}
\frac{\partial \boldsymbol{s}^{l}[t]}{\partial \boldsymbol{u}^{l}[t]} ,\\
\label{eq:observe:temporal gradients errors}
\mathcal{T}^l[t, t^{\prime}] = \frac{\partial\mathcal{L}}{\partial \boldsymbol{s}^{l}[t^{\prime}]} \frac{\partial \boldsymbol{s}^{l}[t^{\prime}]}{\partial \boldsymbol{u}^{l}[t^{\prime}]}
\prod_{t^{\prime\prime}=1}^{t^{\prime}-t} \boldsymbol{\epsilon}^{l}[t^{\prime}-t^{\prime\prime}]  ,
\end{gather}
where the $t^\prime \in [t + 1, T]$. By substituting Eq.\eqref{eq:preliminary:epsilon_after} into Eq.\eqref{eq:observe:temporal gradients errors}, we obtain:
\begin{equation}\label{eq:observe:temporal_gradients_error_part1_part2}
\frac{\partial\mathcal{L}}{\partial \boldsymbol{s}^{l}[t^{\prime}]} 
\frac{\partial \boldsymbol{s}^{l}[t^{\prime}]}{\partial \boldsymbol{u}^{l}[t^{\prime}]}
\underbrace{\gamma^{(t^{\prime}-t)} }_{\text{Part I}}
\prod_{t^{\prime\prime}=1}^{t^{\prime}-t} \underbrace{\left( 1 - V_{\mathrm{th}} \mathbb{H}\left(\boldsymbol{u}^{l}[t^{\prime}-t^{\prime\prime}] \right) \right)  }_{\text{Part II}} .
\end{equation}

The temporal gradient error is a production of two parts.
\textbf{For Part I}, When the difference between $t^\prime$ and $t$ is substantial, Part I gets very close to zero. The $\gamma$ is defined as $1 – \frac{1}{\tau}$, $t$ denotes the time constant $\in [1, +\inf)$. Typically, $\gamma$ is between 0.09 and 0.5. For example, in \cite{meng2023towards, deng2021temporal, xiao2022online} they set $\tau$ = 1.1, 1.3, 1.5, 2.0. For large $t^\prime-t$, $\gamma^{(t^\prime-t)}$ could barely contribute to the $\epsilon$. This could explain our observation in Figure.\ref{fig:obervervation_tau_vs_k}(a).

\textbf{Part II} can be expressed as:
\begin{equation}\label{method: Theoretical Analysis: partII}
1 - V_{\mathrm{th}} \mathbb{H}\left(\boldsymbol{u}^{l}[t]\right) = 
\begin{cases} 
0  & \text{ if  } \frac{1}{2}V_{\mathrm{th}} < \boldsymbol{u}^{l}[t] < \frac{3}{2}V_{\mathrm{th}} \\ 
1  & \text{ otherwise }
\end{cases}.
\end{equation}

More in-depth discussions and proofs of this equation are available in the Appendix.\ref{apx:Detailed Discussion on Temporal Gradient Errors}. From Eq.\eqref{method: Theoretical Analysis: partII}, it can be seen part II has binary values of 0 or 1. More specifically, as long as the neuron fires at least once within the temporal range of $(t+1, T)$, part II = 0 and the corresponding temporal gradient error will vanish and cannot contribute to the backpropagation training process. This is an unavoidable issue in vanilla-LIF models. 

The vanishing of part II is also demonstrated experimentally. For example, in the special case with timestep $T=2$, part I equal to $\gamma$ we could examine the influence of part II and the experiment results are given in Section.\ref{section:Ablation and Analysis}.

To summarize, Eq.\eqref{eq:observe:temporal gradients errors} demonstrates the temporal gradient vanishing due to the vanishing $\epsilon$ in two folds: the multiplication of gamma at large $t^\prime - t$ and the neuron spike between $t^\prime$ and $t$. We define this as the temporal gradient vanishing problem persists with the vanilla-LIF model.

\subsection{The Design of Complementary LIF Model}
To address the temporal gradient errors vanishing problem, we design the Complementary LIF (CLIF) model inspired by biological principles {\color{black}(See detailed in Appendix.\ref{apx:Detailed Discussion on inspired biological principles})}. Besides membrane potential, we introduce another state, termed "Complementary potential". To maintain the efficiency advantage of SNN as well as the broad applicability of our neuron model, our model contains no learnable parameters.

\textit{Decay of Complementary membrane potential}:
Between each timestep, the membrane potential is decayed by $\frac{1}{\tau}\boldsymbol{u}^{l}[t]$. We design our Complementary potential to compensate for the membrane potential decay as follows:
\begin{equation}\label{eq:design:m=m-1*u}
\boldsymbol{m}^l[t] = \boldsymbol{m}^l[t-1] \odot \sigma\left(\frac{1}{\tau}\boldsymbol{u}^l[t] \right) .
\end{equation}

We choose the Sigmoid function as $\sigma$, As $\sigma \in (0,1)$ and the Complementary potential also decays. Nevertheless, the more the membrane potential decays, the less the Complementary potential decays. This design aims to preserve the decayed portion of the membrane potential into Complementary membrane potential.

\textit{Increase of Complementary membrane potential}:
Within each timestep, the Complementary membrane potential is increased by firing
\begin{equation}\label{eq:design:m=m+s}
\boldsymbol{m}^l[t] = \boldsymbol{m}^l[t] + \boldsymbol{s}^l[t].
\end{equation}

If the neuron has fired recently, the membrane potential $\boldsymbol{m}^l[t]$ gets larger. 

\textit{Redesign Reset process:} We revisit the Vanilla LIF model as defined by equation Eq.\eqref{eq:preliminary:ut = ut-1}-\eqref{eq:preliminary:reset}, focusing particularly on LIF’s reset process:
\begin{equation}
\label{eq:design:reset}
\boldsymbol{u}^l[t] = \boldsymbol{u}^l[t] - \boldsymbol{s}^l[t] \odot V_{\mathrm{th}}  .
\end{equation}

The redesigned reset process is given in Eq.\eqref{eq:design:u=u-s(m+vth)}. Compared to the soft reset in Eq.\eqref{eq:design:reset}, each time the neuron fires, the membrane potential is subtracted by another term $\sigma(\boldsymbol{m}^l[t])$ related to the Complementary potential: 
\begin{equation}\label{eq:design:u=u-s(m+vth)}
\boldsymbol{u}^l[t] = \boldsymbol{u}^l[t] - \boldsymbol{s}^l[t] \odot \left( V_{\mathrm{th}} + \sigma(\boldsymbol{m}^l[t]) \right)  .
\end{equation}

If the neuron fires recently, the membrane potential $\boldsymbol{m}^l[t]$ gets larger, and the membrane potential $\boldsymbol{u}^l[t]$ will be subtracted more, suppressing the neuron's firing rate. {\color{black} This mechanism achieves spike frequency adaptation, similar to} the hyper-polarization process in real biological neurons \cite{mccormick1990properties, sanchez2000cellular}. {\color{black} However, unlike classic spike frequency adaptation mechanisms, the adaptation of CLIF depends not only on recently firing activity but also on the recent membrane potential. This means that CLIF can capture more temporal information.}

\textit{Summarizing the above principles, the CLIF model can be derived as following}:
\begin{equation}
\left\{ 
\begin{aligned}
\boldsymbol{u}^l[t] &= (1-\frac{1}{\tau}) \boldsymbol{u}^l[t-1] + \boldsymbol{c}^l[t] ,\\ 
\boldsymbol{s}^l[t] &= \Theta(\boldsymbol{u}^l[t] - V_{\mathrm{th}})  ,\\
\boldsymbol{m}^l[t] &= \boldsymbol{m}^l[t-1] \odot \sigma\left(\frac{1}{\tau} \boldsymbol{u}^l[t]\right) + \boldsymbol{s}^l[t]  ,   \\
\boldsymbol{u}^l[t] &= \boldsymbol{u}^l[t] - \boldsymbol{s}^l[t] \odot \left(V_{\mathrm{th}} + \sigma(\boldsymbol{m}^l[t])\right)   .
\end{aligned}
\right.
\end{equation}
The pseudo-code for the CLIF model is shown in Algorithm.\ref{alg:CLIF}. The simplicity of CLIF is reflected in the fact that we only add two lines of code to LIF neuron model.

\begin{algorithm}[tb]
   \caption{Core function for CLIF model}
   \label{alg:CLIF}
\begin{algorithmic}
\renewcommand{\algorithmiccomment}[1]{\hfill $\triangleright$ #1}
   \STATE {\bfseries Input:} Input $\boldsymbol{c}$, Current Time Step $t$, time constant $\tau$, threshold $V_{\mathrm{th}}$
   \STATE {\bfseries Output:} Spike $\boldsymbol{s}$

   \IF{$t = 0$}
   \STATE Initial $\boldsymbol{u_\mathrm{pre}}$ and $\boldsymbol{m_\mathrm{pre}}$ with all zero
   \ENDIF

   \STATE $\boldsymbol{u} = (1 - \frac{1}{\tau}) \boldsymbol{u_\mathrm{pre}} + \boldsymbol{c} $    \textcolor{blue}{\COMMENT{leaky \& integrate}}
   \STATE $\boldsymbol{s} = \Theta(\boldsymbol{u} - V_{\mathrm{th}})$            \textcolor{blue}{\COMMENT{fire}}
   \STATE $\boldsymbol{m} = \boldsymbol{m_\mathrm{pre}} \odot \sigma\left( \frac{1}{\tau}\boldsymbol{u} \right) + \boldsymbol{s}$  
   \STATE $\boldsymbol{u_\mathrm{pre}} =  \boldsymbol{u} - \boldsymbol{s} \odot \left(V_{\mathrm{th}} + \sigma(\boldsymbol{m}) \right) $            \textcolor{blue}{\COMMENT{reset}}
   \STATE $ \boldsymbol{m_\mathrm{pre}} = \boldsymbol{m}$
   \STATE \textbf{Return} $\boldsymbol{s}$  \textcolor{blue}{\COMMENT{spike output}}
\end{algorithmic}
\end{algorithm}

\subsection{Dynamic and Theoretical Analysis}\label{subsection:Dynamic and Theoretical Analysis}
To validate the effectiveness of the CLIF model, we examine the CLIF model through both case studies and theoretical analysis. 

In the case study, we explore the dynamic properties of both LIF and CLIF models. CLIF features spike frequency adaptation and exhibits a lower firing rate compared to the LIF neuron. This phenomenon is similar to the refractory period or hyperpolarization in the biological neuron \cite{sanchez2000cellular}. More specifically, when the input spikes get dense, the complementary potential gets high, the reset process gets more substantial, as shown in Eq.\eqref{eq:design:u=u-s(m+vth)}. The {\color{black} more detailed dynamic analysis} are illustrated in the Appendix.\ref{apx: The dynamic of CLIF neuron}.

{\color{black}In the theoretical Analysis, we separate the gradient error into spatial and temporal components in Eq.\eqref{eq:method:CLIF_dL_du_analysis}. The details of this derivation are given in Appendix.\ref{apx:The gradients of CLIF model}). This separation demonstrates that CLIF not only contains all temporal gradients in LIF but also contains extra temporal gradient terms. We believe these additional temporal terms contribute to the improved performance of CLIF.}
\begin{equation}\label{eq:method:CLIF_dL_du_analysis}
\begin{aligned}
\frac{\partial\mathcal{L}}{\partial \boldsymbol{u}^{l}[t]} 
= 
{\color{black}
\mathcal{P}^l_{\mathrm{M}}[t]
}
+ &
\sum_{t^{\prime}=t+1}^{T} \mathcal{T}^l_{\mathrm{M1}}[t, t^{\prime}]\\
+ &
\left(\sum_{t^{\prime}=t+1}^{T} \mathcal{T}^l_{\mathrm{M2}}[t, t^\prime] \right) \boldsymbol{\psi}^{l}[t] ,
\end{aligned}
\end{equation}
where $\mathcal{P}^l_{\mathrm{M}}[t]$ and {\color{black}$\mathcal{T}^l_{\mathrm{M}}[t]$ presents the Spatial and Temporal parts of CLIF’s Gradient Errors in respectively. Meanwhile, the M1 and M2 indicate that the Temporal term is divided into two parts.}

{\color{black}Firstly,} the spatial term of CLIF's gradient errors $\mathcal{P}^l_{\mathrm{M}}[t]$ {\color{black}in Eq.\eqref{eq:method:CLIF_dL_du_analysis}} is identical to the counterpart in LIF neuron in Eq.\eqref{eq:observe:spatial gradients errors}. The detailed derivation and proof are given in Appendix.\ref{apx:The gradients of CLIF model}.

{\color{black}Secondly, for the temporal term of CLIF’s gradient errors}, the first temporal part $\mathcal{T}^l_{\mathrm{M1}}[t, t^{\prime}]$ {\color{black}expands as}:
\begin{equation}
\begin{aligned}
\mathcal{T}^l_{\mathrm{M1}}
= &
\left(
{\color{black}
\frac{\partial\mathcal{L}}{\partial \boldsymbol{s}^{l}[t^{\prime}]} 
\frac{\partial\boldsymbol{s}^{l}[t^{\prime}]}{\partial \boldsymbol{u}^{l}[t^{\prime }]}
}
+
\frac{\partial\mathcal{L}}{\partial \boldsymbol{m}^{l}[t^{\prime}]} \boldsymbol{\psi}^{l}[t^{\prime}] 
\right) \prod_{t^{\prime\prime}=1}^{t^{\prime}-t} 
\boldsymbol{\xi}^{l}[t^{\prime}-t^{\prime\prime}] ,
\end{aligned}
\end{equation}
where the $\boldsymbol{\xi}^{l}[t]$ is defined as:
\begin{equation}
\begin{aligned}
\boldsymbol{\xi}^{l}[t] 
= &
{\color{black}\boldsymbol{\epsilon}^{l}[t]} 
+  
\frac{\partial \boldsymbol{u}^{l}[t+1]}{\partial \boldsymbol{m}^{l}[t]} \boldsymbol{\psi}^{l}[t],
\end{aligned}
\end{equation}
this term can be simplified to a product involving the constant term $\gamma$, the same as Eq.\eqref{eq:observe:temporal_gradients_error_part1_part2}. The issue discussed in Part I of Section.\ref{section: Limitation of LIF-based SNN with SG}, regarding the vanishing of temporal gradients, also applies here. Where $\boldsymbol{\psi}^{l}[t]$ is non-negative (see Appendix.\ref{apx:The gradients of CLIF model}). $\boldsymbol{\psi}^{l}[t]$ is defined as:
\begin{equation}
\boldsymbol{\psi}^{l}[t] 
\equiv 
\frac{\partial \boldsymbol{m}^{l}[t]}{\partial \boldsymbol{u}^{l}[t]} +
\frac{\partial \boldsymbol{m}^{l}[t]}{\partial \boldsymbol{s}^{l}[t]} 
\frac{\partial \boldsymbol{s}^{l}[t]}{\partial \boldsymbol{u}^{l}[t]} .
\end{equation}

{\color{black}Finally}, for the other {\color{black}temporal} item of CLIF's gradient errors, $\mathcal{T}^l_{\mathrm{M2}}[t, t^{\prime}]$, can be expressed as:
\begin{equation}
\begin{aligned}
\mathcal{T}^l_{\mathrm{M2}}[t, t^{\prime}] = &
\frac{\partial\mathcal{L}}{\partial \boldsymbol{u}^{l}[t^{\prime}]} 
\frac{\partial \boldsymbol{u}^{l}[t^{\prime}]}{\partial \boldsymbol{m}^{l}[t^{\prime}-1]} 
\prod_{t^{\prime\prime}=2}^{t^{\prime}-t} \boldsymbol{\rho}^{l}[t^{\prime} - t^{\prime\prime}] .
\end{aligned}
\end{equation}

\begin{figure}[hbt]
\label{fig:training process}
\centering
	\subfigure[Exchange neuron] {\includegraphics[width=.22 \textwidth]{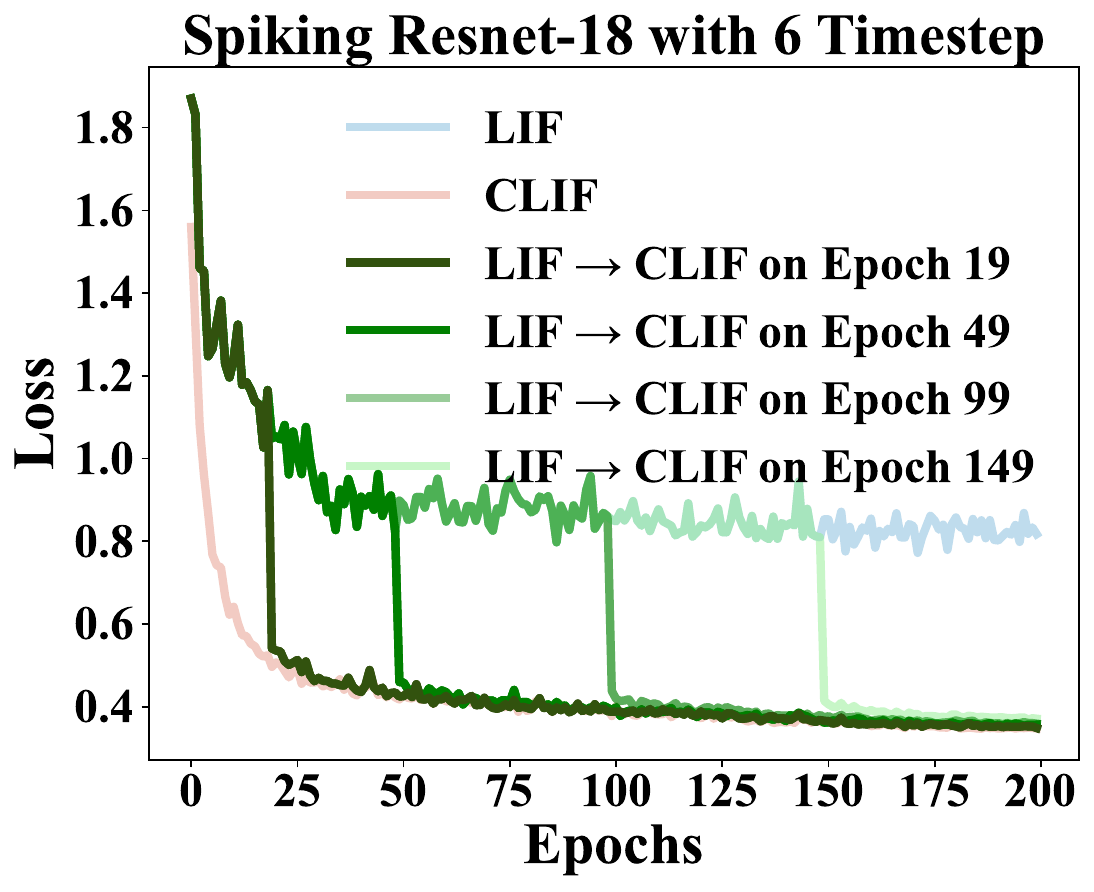}}
 	\subfigure[The timestep ablation] {\includegraphics[width=.22 \textwidth]{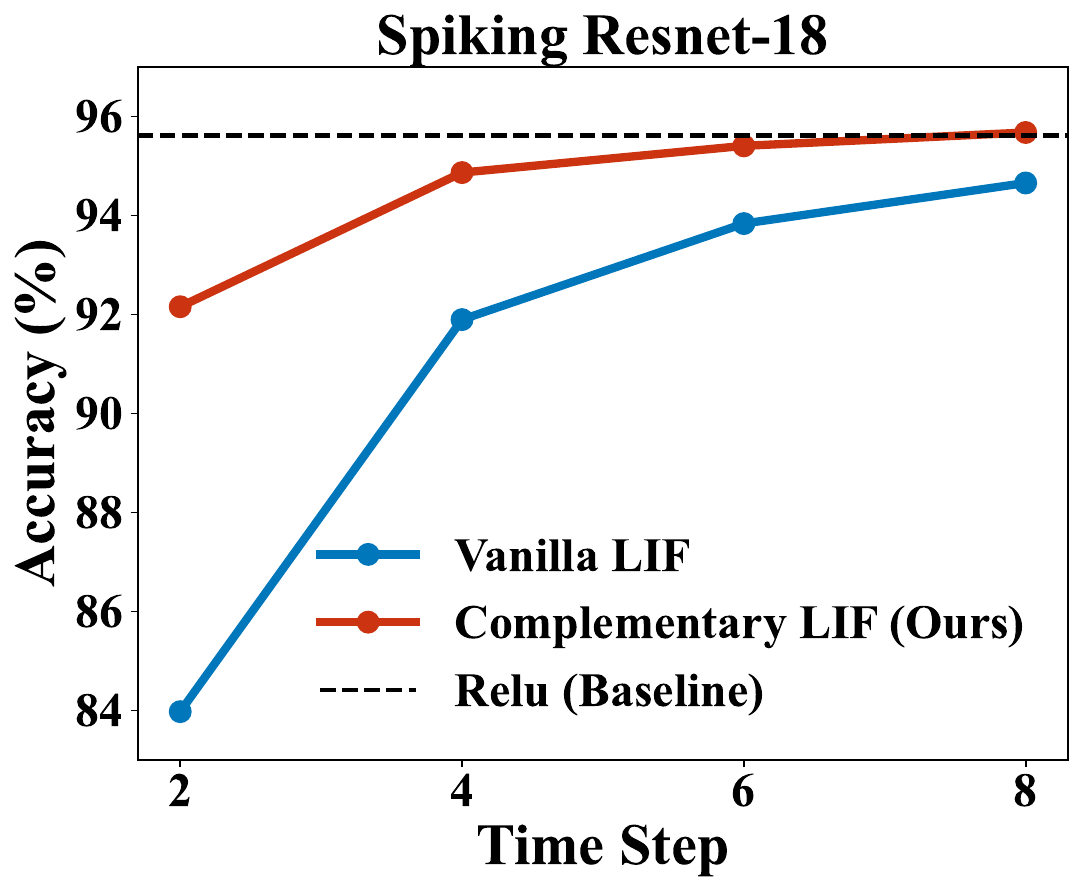}}
	\caption{(a) Loss function vs epochs. Each color presents a case of either LIF, CLIF, or exchanging from LIF to CLIF at a given epoch during training. (b) Comparison of the accuracy of LIF and CLIF at various timestep. Both experiments are evaluated on the CIFAR10 task with Spiking ResNet-18.}
	\label{fig:acc_vs_T}
\end{figure}

This term indicates that additional Temporal Gradient Errors components are generated. {\color{black} We believe this additional part contributes to the performance improvements. As $\mathcal{T}^l_{\mathrm{M2}}$ does not decay as fast as $\prod \xi$ over timestep, this phenomenon could be observed in dynamic analysis in the Appendix.\ref{apx: The dynamic of CLIF neuron}. Therefore, this item of gradient errors contributes to compensate for the vanishing of temporal gradients in LIF, leading to a further reduction in the loss. This assertion is also verifiable in Figure.\ref{fig:acc_vs_T}(a) in the Experiment.
}

\begin{figure}[hbt]
\label{fig:neuron ablation}
	\centering
 	\subfigure[Cifar10 (Timestep=8)] {\includegraphics[width=.218 \textwidth]{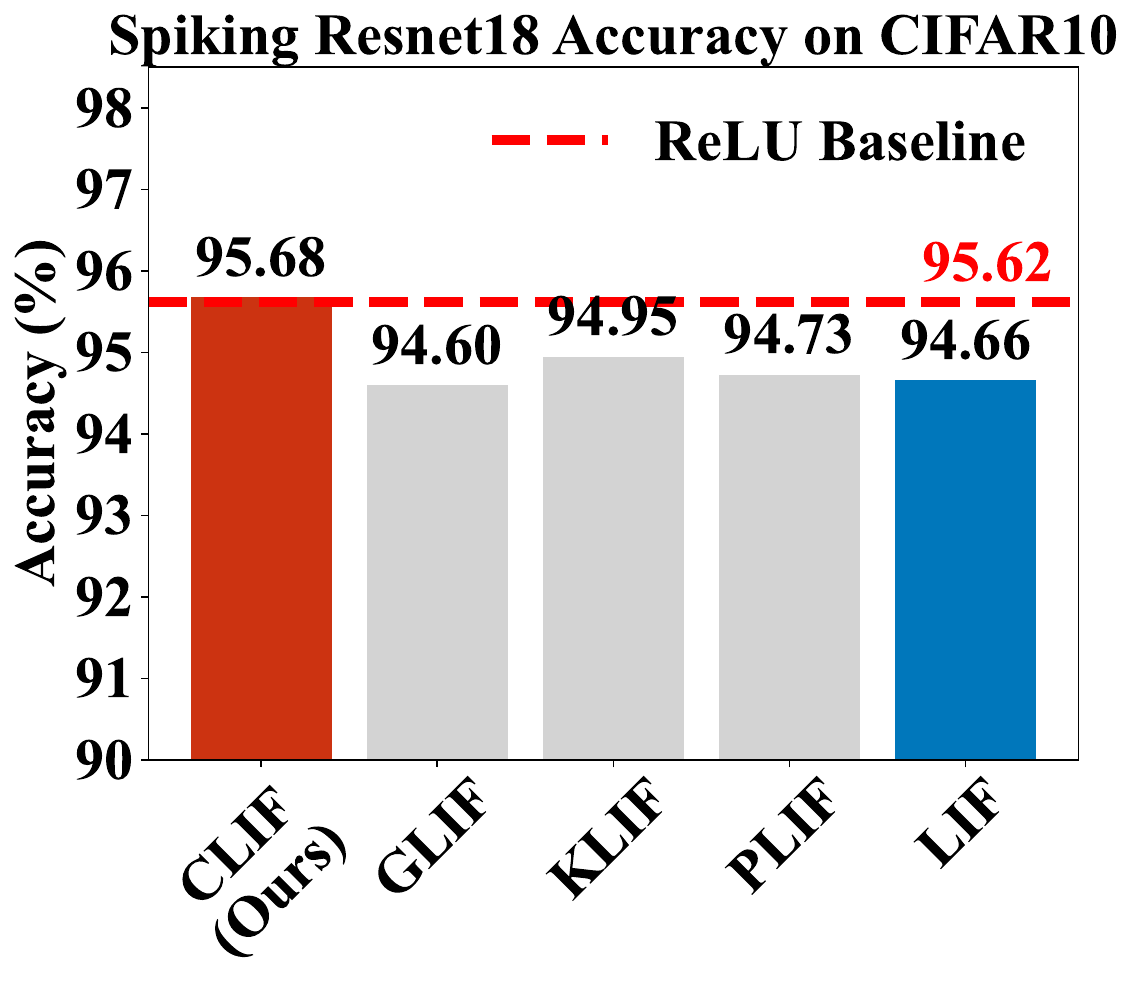}}
	\subfigure[Cifar100 (Timestep=6)] {\includegraphics[width=.22 \textwidth]{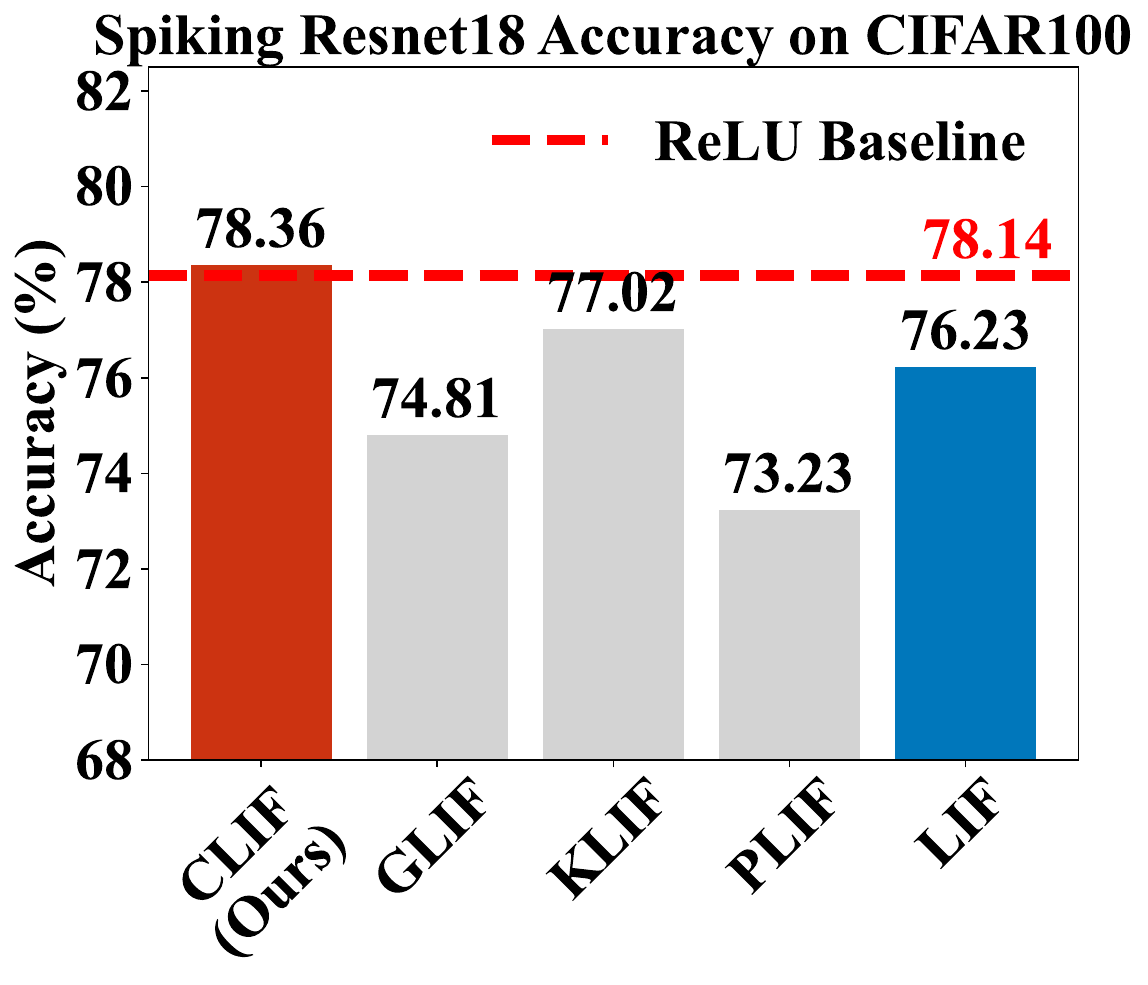}}
	\caption{Comparative accuracy of Spiking ResNet-18. Panels (a) CIFAR10 using 8 timestep (b) CIFAR100 using 6 timestep with different neuron. }
	\label{fig:different neuron}
\end{figure}

\begin{table*}[t]
\centering
\caption{Comparing the state-of-the-art methods on static image datasets. The asterisk $(*)$ indicates the utilization of data augmentation strategies, including auto-augmentation and/or CutMix, our implementation directly following \cite{fang2024parallel}. The implemented ReLU-based ANN shares identical structures and hyper-parameters with SNN.
}
\label{tab:static_img_results}
\begin{center}
\begin{small}
\renewcommand{\arraystretch}{1.2}
\begin{tabular}{cccccc}
\toprule
\textbf{Dataset} & \textbf{Method} & \textbf{Spiking Network} & \textbf{Neuron Model} & \textbf{Timestep} & \textbf{Accuracy(\%)} \\
\midrule
\multirow{12}{*}{
\rotatebox[origin=c]{90}{\textbf{CIFAR-10}}}   
             & ANN / $\text{ANN}^{\text{*}}$ & ResNet-18 & ReLU & 1 & 95.62 / 96.65 \\
            \cline{2-6} 
             & Dspike \cite{li2021differentiable}$^{\textit{'NIPS}}$ & Modified ResNet-18 & LIF & 6 & 93.50 \\ 
             & 
               PLIF \cite{fang2021incorporating}$^{\textit{'ICCV}}$ & PLIF Net & PLIF & 6 & 94.25 \\ 
             & DSR \cite{meng2022training}$^{\textit{'CVPR}}$ & RreAct-ResNet-18 & LIF & 20 & 95.40 \\ 
             & 
               GLIF \cite{yao2022glif}$^{\textit{'NIPS}}$ & ResNet-18 & GLIF & 6 & 94.88 \\ 
             & 
               KLIF \cite{jiang2023klif}$^{\textit{'ArXiv}}$ & Modified PLIF Net & KLIF & 10 & 92.52 \\ 
             & 
               $\text{PSN}^{\text{*}}$ \cite{fang2024parallel}$^{\textit{'NIPS}}$ & Modified PLIF Net & PSN & 4 & 95.32 \\ 
             & SML \cite{deng2023surrogate}$^{\textit{'ICML}}$ & ResNet-18 & LIF & 6 & 95.12 \\
             & $\text{ASGL}^{\text{*}}$  \cite{wang2023adaptive}$^{\textit{'ICML}}$ & ResNet-18 & LIF & 4 & 95.35 \\
             \cline{2-6} 
             & \multirow{3}{*}{\textbf{Ours} / $\textbf{Ours}^{\text{*}}$ } & \multirow{3}{*}{\textbf{ResNet-18}} & \multirow{3}{*}{\textbf{CLIF}} & \textbf{4} & \textbf{94.89} / \textbf{96.01}  \\ 
                    &   &  &   & \textbf{6} & \textbf{95.41} / \textbf{96.45}\\  
                    &   &  &   & \textbf{8} & \textbf{95.68} / \textbf{96.69}\\     
\midrule
\multirow{9}{*}{
\rotatebox[origin=c]{90}{\textbf{CIFAR-100}} }  
            & ANN / $\text{ANN}^{\text{*}}$  & ResNet-18 & ReLU & 1 & 78.14 / 80.89\\ \cline{2-6} 
            & Dspike \cite{li2021differentiable}$^{\textit{'NIPS}}$ & Modified ResNet-18 & LIF & 6 & 74.24 \\ 
            & DSR \cite{meng2022training}$^{\textit{'CVPR}}$ & RreAct-ResNet-18 & LIF & 20 & 78.50 \\
            & 
              GLIF \cite{yao2022glif}$^{\textit{'NIPS}}$ & ResNet-18 & GLIF & 6 & 77.28 \\
            & SML \cite{deng2023surrogate}$^{\textit{'ICML}}$ & ResNet-18 & LIF & 6 & 78.00 \\
            & $\text{ASGL}^{\text{*}}$  \cite{wang2023adaptive}$^{\textit{'ICML}}$ & ResNet-18 & LIF & 4 & 77.74 \\
            \cline{2-6} 
            & \multirow{3}{*}{\textbf{Ours} / $\textbf{Ours}^{\text{*}}$ } & \multirow{3}{*}{\textbf{ResNet-18}} & \multirow{3}{*}{\textbf{CLIF}} & \textbf{4} & \textbf{77.00} / \textbf{79.69} \\
            &   &   &   & \textbf{6} & \textbf{78.36} / \textbf{80.58} \\     
            &   &   &   & \textbf{8} & \textbf{78.99} / \textbf{80.89}\\    
\midrule    
\multirow{6}{*}{
\rotatebox[origin=c]{90}{\textbf{Tiny-ImageNet}} }  
            & $\text{ANN}$ & VGG-13 & ReLU & 1 & 59.77 \\
            \cline{2-6}
            & Online LTL \cite{yang2022training}$^{\textit{'NIPS}}$ & VGG-13 & LIF & 6 & 55.37 \\
            & Joint A-SNN \cite{guo2023joint}$^{\textit{'Pattern Recognit}}$ & VGG-16 & LIF & 4 & 55.39 \\
            & $\text{ASGL}$  \cite{wang2023adaptive}$^{\textit{'ICML}}$ & VGG-13 & LIF & 8 & 56.81 \\
            \cline{2-6} 
            & \multirow{2}{*}{$\textbf{Ours}$}& \multirow{2}{*}{\textbf{VGG-13}} & \multirow{2}{*}{\textbf{CLIF}} & \textbf{4} & \textbf{63.16} \\
            &   &   &    & \textbf{6} & \textbf{64.13} \\
\bottomrule
\end{tabular}
\end{small}
\end{center}
\end{table*}

\begin{table*}[t]
\centering
\caption{Comparing the SOTA neuronal models by using neuromorphic datasets. The footnote in the table indicates implementation directly in open source code by only modifying neurons: $^1$\cite{yao2024spike}, $^2$\cite{fang2024parallel} with data augmentation.  'T' denotes the number of the timestep employed.}
\label{tab:experiment:neuromorchic datasets}
\begin{small}
\begin{center}
\renewcommand{\arraystretch}{1.2}
\begin{tabular}{cccccc}
\toprule
\textbf{Dataset} & \textbf{Method} & \textbf{Spiking Network} & \textbf{Neuron Model} & \textbf{T} & \textbf{Accuracy(\%)} \\
\midrule
\multirow{6}{*}{
\textbf{DVS-Gesture}}
             & PLIF \cite{fang2021incorporating}$^{\textit{'ICCV}}$  & PLIF Net & PLIF & 20 & 97.57 \\
             & KLIF \cite{jiang2023klif}$^{\textit{'ArXiv}}$ & Modified PLIF Net & KLIF & 12 & 94.10 \\ 
             \cline{2-6} 
             & \multirow{4}{*}{\textbf{Ours}} & \multirow{2}{*}{\textbf{Spiking-Vgg11}} &  LIF & \multirow{2}{*}{20} & 97.57 \\  
             &   &   &  \textbf{CLIF}  &  & \textbf{97.92} \\  
                  \cline{3-6} 
              &   & \multirow{2}{*}{\textbf{Spike-Driven-Transformer}$^1$}  &  LIF  &  \multirow{2}{*}{16} & 98.26 \\
             &   &    &  \textbf{CLIF}  &   & \textbf{99.31} \\
             \hline
\multirow{9}{*}{
\textbf{CIFAR10-DVS}}  
             & PLIF \cite{fang2021incorporating}$^{\textit{'ICCV}}$ & PLIF Net & PIF & 20 & 74.80 \\ 
             & KLIF \cite{jiang2023klif}$^{\textit{'ArXiv}}$ & Modified PLIF Net & KLIF & 15 & 70.90 \\ 
             & GLIF \cite{yao2022glif}$^{\textit{'NIPS}}$ & 7B-wideNet & GLIF & 16 & 78.10 \\ 
             & PSN \cite{fang2024parallel}$^{\textit{'NIPS}}$ & VGGSNN & PSN & 10 & 85.90 \\ 
             \cline{2-6}  
             &  \multirow{4}{*}{\textbf{Ours}}  &  \multirow{2}{*}{\textbf{Spiking-Vgg11}} 
                        & LIF & \multirow{2}{*}{16} & 78.05 \\   
             &    &    & \textbf{CLIF} &  & \textbf{79.00} \\     
             \cline{3-6}
                 &     &    \multirow{2}{*}{\textbf{VGGSNN}$^2$}  
                       & \multirow{1}{*}{LIF} & \multirow{2}{*}{10} & 84.90 \\ 
            &      &         &  \textbf{CLIF} &   & \textbf{86.10} \\   
\bottomrule
\end{tabular}
\end{center}
\end{small}
\end{table*}


\section{Experiment}
To validate the effectiveness of the proposed CLIF neuron, we conduct a set of ablation studies. These studies are designed to evaluate the underlying principles of the CLIF model, to examine the effect of various timestep, and to conduct comparative analyses between the CLIF model and other neuron models. Following the ablation study, we extend our experiments to cover a diverse range of data types, including static image datasets and neuromorphic event-based datasets. Details on the experimental implementation are provided in the Appendix.\ref{appendix:Dataset Description and Preprocessing}.

\subsection{Ablation and Analysis}\label{section:Ablation and Analysis}
We conduct two experiments to compare LIF and CLIF: the loss of CLIF versus LIF via training epochs, and the accuracy of CLIF versus LIF via timestep. For a fair comparison, except for the control variable, the same optimizer setting, random seed, architecture, loss function, weight initialization and all hyperparameters are employed.

\textbf{Exchange of Neuron Models} 
In the first ablation study, we use the Spiking Resnet18 with 6 timestep by BPTT training. The training of the network begins with LIF neuron and later transitions to CLIF neurons at a designated epoch. As shown in Figure.\ref{fig:acc_vs_T}(a), a few epochs after Exchange to CLIF, the loss decreases significantly compared to LIF. Moreover, the decay of loss over training epochs is much faster when training with CLIF than LIF.

We extend the loss comparison to various tasks and network backbones (see Appendix.\ref{appendix:The Loss Comparing between LIF and CLIF}). In all experiments, CLIF neuron’s loss converges faster than LIF’s, the converged loss is also lower. As such, one can conclude that CLIF neurons are more effective in capturing error information both precisely and efficiently, suggesting the higher training accuracy and efficiency of CLIF.

\textbf{Temporal Ablation}
To demonstrate CLIF’s efficacy in capturing temporal gradient errors over longer period, we verify the performance comparison between CLIF and LIF with various timestep. The ablation study is conducted in CIFAR10 task with Resnet-18 as the backbone network. The results are illustrated in Figure.\ref{fig:acc_vs_T}(b). When the timestep is 1, CLIF and LIF cannot make any temporal gradient over membrane potential and yield identical accuracy of 92.7\% (not shown in the figure). At higher timestep, CLIF always outperforms LIF neuron in accuracy. Specifically, at the timestep of 2 the vanilla LIF model encounters the problem of temporal gradient vanishing, as detailed in Part II of the theoretical analysis (Section 4.1). Remarkably, even with just two timestep, the performance of CLIF is already significantly better than LIF, this also verifies that CLIF can better capture the temporal gradient error information.

\subsection{Comparison and Quantitative analysis}
We conduct two sets of comparison experiments to ascertain the effectiveness of CLIF: comparison with Different Neurons, and comparison with SOTA methods.

\textbf{Comparison with Different Neuron} In order to verify whether CLIF is more effective than existing methods, we self-implement and compare vanilla-LIF \cite{wu2018spatio}, PLIF \cite{fang2021incorporating}, KLIF \cite{jiang2023klif} and GLIF \cite{yao2022glif}. Except for the neuron models, all other experimental conditions are kept identical, including the backbone architecture, random seeds and hyperparameters. CLIF exhibits superior performance over other neuron benchmarks in the CIFAR10 and CIFAR100 datasets, as shown in Figure.\ref{fig:different neuron}(a) and (b). {\color{black} PLIF and GLIF include additional training parameters, so additional hyperparameters tuning and more training epochs are required to converge.} Moreover, CLIF can achieve slightly better performance with ReLU-based ANNs.

\textbf{Comparison with SOTA methods}
We compare our approach with state-of-the-art methods in two categories of datasets: static dataset (CIFAR10/100 and Tiny-ImageNet), as summarized in Table.\ref{tab:static_img_results} and neuromorphic dataset (DVS-Gesture and CIFAR10-DVS), as summarized in Table.\ref{tab:experiment:neuromorchic datasets}. We not only explore the diversity of datasets but also the diversity in network backbone, including ResNet, VGG and Transformer. We also compare the fire rate and energy consumption of LIF, CLIF and ReLU. In short, CLIF has lower fire rate and similar energy consumption as LIF. Detailed statistics of fire rate and power consumption are described in the Appendix.\ref{appendix:The Comparing of Computing Efficiency}.

\textbf{Static Image Datasets} Table.\ref{tab:static_img_results}, reveals that CLIF always outperforms its LIF counterpart and surpasses all other SNNs neuron models within the same network backbone. CLIF achieves 96.69\% accuracy on CIFAR-10 and 80.89\% on CIFAR-100 datasets, not only outperforming other SNN models but also slightly outperforming ReLU-based ANNs counterpart. 
In Tiny-ImageNet, CLIF achieves 64.13\% accuracy with 6 timestep, significantly better than the other SNNs and ANN within the same network backbone. These results demonstrate CLIF's competitiveness with existing neuron models.

\textbf{Neuromorphic Datasets} To validate that our method can handle spatio-temporal error backpropagation properly, we conduct experiments on different neuromorphic datasets of DVS-Gesture \cite{amir2017low} and DVSCIFAR10 \cite{li2017cifar10}. The results are summarized in Table 3. For DVS Gesture, CLIF accuracy is 97.92\% with Spiking-VGG11 as backbone and 99.31\% with Spike-Driven Transformer \cite{yao2024spike} as the backbone, surpassing LIF based model by 0.35\% and 1.05\%, respectively. On the DVSCIFAR10 dataset, CLIF accuracy is 79.00\% with Spiking-VGG11 as the backbone and 86.10\% with VGGSNN as the backbone, surpassing LIF based model by 0.95\% and 1.20\%, respectively. It is worth noting that CLIF features the highest accuracy of 86.10\% in all methods in this dataset. This is achieved by simply replacing the neuron model with CLIF in the network architecture.

\section{Conclusion}
In this work, we investigate the vanishing of temporal gradient effort and propose the CLIF model with richer temporal gradient. CLIF features Complementary membrane potential on top of the conventional membrane potential and creates extra paths in temporal gradient computation while keeping binary output. CLIF shows experimentally clear performance advantage over other neuron models in various tasks with different network backbones. Moreover, CLIF achieves comparable performance with ANNs with identical architecture and hyperparameters. Furthermore, the CLIF model is characterized by its generalizability and versatility, it can apply to various backbones, and it’s interchangeable with vanilla LIF neuron. Nevertheless, due to the mathematical complexity of the CLIF’s neuron equations, a more thorough analysis of the temporal gradient errors and the neuron’s dynamic behavior remains to be performed in the future.

\section*{Acknowledgments}
This work was supported in part by the Young Scientists Fund of the National Natural Science Foundation of China (Grant 62305278), by the Guangzhou Municipal Science and Technology Project under Grant 2023A03J0013 and Grant 2024A04J4535.

\section*{Impact Statement}
This paper presents work whose goal is to advance the field of Spiking Neural Networks and Neuromorphic Computing. There are many potential societal consequences of our work, none which we feel must be specifically highlighted here.

\nocite{langley00}

\bibliography{example_paper}
\bibliographystyle{icml2024}

\onecolumn
\appendix
\newpage

\section{Notation in the Paper}\label{apx: Notation in the Paper}
Throughout the paper and this Appendix, we use the following notations, which mainly follow this work\cite{wang2023adaptive}. We follow the conventions representing vectors and matrices with bold italic letters and bold capital letters respectively, such as $\boldsymbol{s}$ and $\boldsymbol{W}$. For this symbol $\boldsymbol{W}^{\top}$ represents transposing the matrix. For a function $f (\boldsymbol{x}) : \mathbb{R}^{d_1} \rightarrow \mathbb{R}^{d_2}$, we use $\nabla_{\boldsymbol{x}} f$ instead of $\frac{\partial f}{\partial \boldsymbol{x}}$ to represent the $1^{\mathrm{th}}$ derivative gradients of $f$ with respect to the variable $x$ in the absence of ambiguity. For two vectors $\boldsymbol{u_1}$ and $\boldsymbol{u_2}$, we use $\boldsymbol{u_1} \odot \boldsymbol{u_2}$ to represent the element-wise product.

\section{LIF-Based BPTT with Surrogate Gradient}\label{apx: LIF-based BPTT with surrogate gradient}
This section is mainly referenced from \cite{wu2018spatio, meng2023towards}. Firstly, we recall the LIF model Eq.\eqref{eq:preliminary:ut = ut-1} - \eqref{eq:preliminary:reset}. We then rewrite the LIF model with soft reset mechanism:

\begin{equation}
\boldsymbol{u}^{l}[t] = (1 - \frac{1}{\tau}) \left( \boldsymbol{u}^l[t - 1] - V_{\mathrm{th}} \boldsymbol{s}^l[t-1] \right)  + \boldsymbol{W}^l \boldsymbol{s}^{l-1}[t], 
\end{equation}

\begin{equation}
s^l[t]  = \Theta (\boldsymbol{u}^{l}[t] - V_{\mathrm{th}})
= 
\begin{cases}
1, & \text{if } \boldsymbol{u}^{l}[t] \geq V_{\mathrm{th}} \\
0, & \text{otherwise}
\end{cases} 
\end{equation}

$\gamma$ is defined as $\gamma \equiv 1 - \frac{1}{\tau} $, then we recall the gradient in Eq.\eqref{eq:preliminary:dL_dW}-\eqref{eq:preliminary:eligibility}:
\begin{equation}\label{eq:appdix:dL_dW}
\nabla_{\boldsymbol{W}^l}\mathcal{L}=\sum_{t=1}^T \frac{\partial\mathcal{L}}{\partial \boldsymbol{u}^{l}[t]} \frac{\partial \boldsymbol{u}^{l}[t]}{\partial \boldsymbol{W}^{l}} ,l=\mathrm{L},\mathrm{L}-1,\cdots,1  ,
\end{equation}
where $\mathcal{L}$ represents the loss function. 
For the left part we recursively evaluate:
\begin{equation}\label{eq:appdix:dL_dut}
\frac{\partial\mathcal{L}}{\partial \boldsymbol{u}^{l}[t]} 
= \frac{\partial\mathcal{L}}{\partial \boldsymbol{s}^{l}[t]} \frac{\partial  \boldsymbol{s}^{l}[t]}{\partial \boldsymbol{u}^{l}[t]} 
+ \frac{\partial\mathcal{L}}{\partial \boldsymbol{u}^{l}[t+1]} 
\boldsymbol{\epsilon}^{l}[t] ,
\end{equation}
where $\boldsymbol{\epsilon}^{l}[t]$ for LIF model can be defined as follows: 
\begin{equation}\label{eq:appdix:eligibility}
\boldsymbol{\epsilon}^{l}[t] \equiv \frac{\partial \boldsymbol{u}^{l}[t+1]}{\partial \boldsymbol{u}^{l}[t]} + \frac{\partial \boldsymbol{u}^{l}[t+1]}{\partial \boldsymbol{s}^{l}[t]} \frac{\partial \boldsymbol{s}^{l}[t]}{\partial \boldsymbol{u}^{l}[t]},
\end{equation}

Proof of the Eq.\eqref{eq:preliminary:dL_dU} and Eq.\eqref{eq:preliminary:dL_dS}.
\begin{proof}
Firstly, we only consider the effect of the temporal dimension in Eq.\eqref{eq:appdix:dL_dut}.
When $t = T $, where Eq.\eqref{eq:appdix:dL_dut} deduce as:
\begin{equation}\label{eq:appdix:dL_dU_T}
\frac{\partial\mathcal{L}}{\partial \boldsymbol{u}^{l}[T]} 
= \frac{\partial\mathcal{L}}{\partial \boldsymbol{s}^{l}[T]} \frac{\partial  \boldsymbol{s}^{l}[T]}{\partial \boldsymbol{u}^{l}[T]}.
\end{equation}

When $1 \leq t < T $, with the chain rule, the Eq.\eqref{eq:appdix:dL_dut} can be further calculated recursively:
\begin{equation}\label{eq:appdix:dL_dut_deduce}
\begin{aligned}
\frac{\partial\mathcal{L}}{\partial \boldsymbol{u}^{l}[t]} 
=& \frac{\partial\mathcal{L}}{\partial \boldsymbol{s}^{l}[t]} \frac{\partial  \boldsymbol{s}^{l}[t]}{\partial \boldsymbol{u}^{l}[t]} 
+ \frac{\partial\mathcal{L}}{\partial \boldsymbol{u}^{l}[t+1]} 
\boldsymbol{\epsilon}^{l}[t] \\
=& \frac{\partial\mathcal{L}}{\partial \boldsymbol{s}^{l}[t]} \frac{\partial  \boldsymbol{s}^{l}[t]}{\partial \boldsymbol{u}^{l}[t]} 
+ \underbrace{
\left( 
\frac{\partial\mathcal{L}}{\partial \boldsymbol{s}^{l}[t+1]} \frac{\partial  \boldsymbol{s}^{l}[t+1]}{\partial \boldsymbol{u}^{l}[t+1]} 
+ \frac{\partial\mathcal{L}}{\partial \boldsymbol{u}^{l}[t+2]} 
\boldsymbol{\epsilon}^{l}[t+1] 
\right)}_{\mathrm{expansion}} 
\boldsymbol{\epsilon}^{l}[t] \\
=& \frac{\partial\mathcal{L}}{\partial \boldsymbol{s}^{l}[t]} \frac{\partial  \boldsymbol{s}^{l}[t]}{\partial \boldsymbol{u}^{l}[t]} 
+ 
\frac{\partial\mathcal{L}}{\partial \boldsymbol{s}^{l}[t+1]} \frac{\partial  \boldsymbol{s}^{l}[t+1]}{\partial \boldsymbol{u}^{l}[t+1]} \boldsymbol{\epsilon}^{l}[t] 
+ \frac{\partial\mathcal{L}}{\partial \boldsymbol{u}^{l}[t+2]} 
\boldsymbol{\epsilon}^{l}[t+1] \boldsymbol{\epsilon}^{l}[t] \\
=& \frac{\partial\mathcal{L}}{\partial \boldsymbol{s}^{l}[t]} \frac{\partial  \boldsymbol{s}^{l}[t]}{\partial \boldsymbol{u}^{l}[t]} 
+ 
\frac{\partial\mathcal{L}}{\partial \boldsymbol{s}^{l}[t+1]} \frac{\partial  \boldsymbol{s}^{l}[t+1]}{\partial \boldsymbol{u}^{l}[t+1]} \boldsymbol{\epsilon}^{l}[t] 
+ 
\underbrace{
\left( 
\frac{\partial\mathcal{L}}{\partial \boldsymbol{s}^{l}[t+2]} \frac{\partial  \boldsymbol{s}^{l}[t+2]}{\partial \boldsymbol{u}^{l}[t+2]} 
+ \frac{\partial\mathcal{L}}{\partial \boldsymbol{u}^{l}[t+3]} 
\boldsymbol{\epsilon}^{l}[t+2] 
\right)}_{\mathrm{expansion}}
\boldsymbol{\epsilon}^{l}[t+1] \boldsymbol{\epsilon}^{l}[t] \\
=& \frac{\partial\mathcal{L}}{\partial \boldsymbol{s}^{l}[t]} \frac{\partial  \boldsymbol{s}^{l}[t]}{\partial \boldsymbol{u}^{l}[t]} 
+ 
\frac{\partial\mathcal{L}}{\partial \boldsymbol{s}^{l}[t+1]} \frac{\partial \boldsymbol{s}^{l}[t+1]}{\partial \boldsymbol{u}^{l}[t+1]} \boldsymbol{\epsilon}^{l}[t] 
+ 
\frac{\partial\mathcal{L}}{\partial \boldsymbol{s}^{l}[t+2]} \frac{\partial  \boldsymbol{s}^{l}[t+2]}{\partial \boldsymbol{u}^{l}[t+2]} \boldsymbol{\epsilon}^{l}[t+1] \boldsymbol{\epsilon}^{l}[t] \\
& + \frac{\partial\mathcal{L}}{\partial \boldsymbol{u}^{l}[t+3]} 
\boldsymbol{\epsilon}^{l}[t+2] 
\boldsymbol{\epsilon}^{l}[t+1] \boldsymbol{\epsilon}^{l}[t] \\
=& ...... \\
=& \frac{\partial\mathcal{L}}{\partial \boldsymbol{s}^{l}[t]} \frac{\partial  \boldsymbol{s}^{l}[t]}{\partial \boldsymbol{u}^{l}[t]} 
+ 
\frac{\partial\mathcal{L}}{\partial \boldsymbol{s}^{l}[t+1]} \frac{\partial \boldsymbol{s}^{l}[t+1]}{\partial \boldsymbol{u}^{l}[t+1]} \boldsymbol{\epsilon}^{l}[t] 
+ 
\frac{\partial\mathcal{L}}{\partial \boldsymbol{s}^{l}[t+2]} \frac{\partial  \boldsymbol{s}^{l}[t+2]}{\partial \boldsymbol{u}^{l}[t+2]} \boldsymbol{\epsilon}^{l}[t+1] \boldsymbol{\epsilon}^{l}[t] \\
&+ ... + \frac{\partial\mathcal{L}}{\partial \boldsymbol{s}^{l}[T]} \frac{\partial \boldsymbol{s}^{l}[T]}{\partial \boldsymbol{u}^{l}[T]}
\boldsymbol{\epsilon}^{l}[T-1] 
\boldsymbol{\epsilon}^{l}[T-2] ... \boldsymbol{\epsilon}^{l}[t] 
\end{aligned},
\end{equation}
after iterative expansion, we can inductively summarize the Eq.\eqref{eq:appdix:dL_dU_T} and \eqref{eq:appdix:dL_dut_deduce} to obtain this formula Eq.\eqref{eq:appdix:1<=dL_dU_t<=T}:
\begin{equation}\label{eq:appdix:1<=dL_dU_t<=T}
\begin{aligned}
\frac{\partial\mathcal{L}}{\partial \boldsymbol{u}^{l}[t]}  = \frac{\partial\mathcal{L}}{\partial \boldsymbol{s}^{l}[t]}  \frac{\partial \boldsymbol{s}^{l}[t]}{\partial \boldsymbol{u}^{l}[t]}
+ \sum_{t^{\prime}=t+1}^{T}
\frac{\partial\mathcal{L}}{\partial \boldsymbol{s}^{l}[t^{\prime}]} \frac{\partial \boldsymbol{s}^{l}[t^{\prime}]}{\partial \boldsymbol{u}^{l}[t^{\prime }]}
\prod_{t^{\prime\prime}=1}^{t^{\prime}-t} \boldsymbol{\epsilon}^{l}[t^{\prime}-t^{\prime\prime}] .
\end{aligned}
\end{equation}

Secondly, under the $t \in [1,T]$ situation, we consider the different layer. For last layer, we substitute $l = L$ into Eq.\eqref{eq:appdix:1<=dL_dU_t<=T} as:
\begin{equation}\label{eq:appdix:dL_du_Llayer}
\begin{aligned}
\frac{\partial\mathcal{L}}{\partial \boldsymbol{u}^{L}[t]} 
= \frac{\partial\mathcal{L}}{\partial \boldsymbol{s}^{L}[t]} \frac{\partial \boldsymbol{s}^{L}[t]}{\partial \boldsymbol{u}^{L}[t]}
+\sum_{t^{\prime}=t+1}^T \frac{\partial\mathcal{L}}{\partial \boldsymbol{s}^{L}[t^{\prime}]} \frac{\partial \boldsymbol{s}^{L}[t^{\prime}]} {\partial \boldsymbol{u}^{L}[t^{\prime}]}\prod_{t^{\prime\prime}=1}^{t^{\prime}-t}\epsilon^{L}[t^{\prime}-t^{\prime\prime}].
\end{aligned}
\end{equation}
 
For the intermediate layer $l = L-1, ... , 1$, according to the chain rule, the $\frac{\partial\mathcal{L}}{\partial \boldsymbol{u}^{l}[t]} $ can be obtained from the previous layer $\frac{\partial\mathcal{L}}{\partial \boldsymbol{u}^{l+1}[t]} $, the Eq.\eqref{eq:appdix:1<=dL_dU_t<=T} can be shown in:
\begin{equation}\label{eq:appendix:dL_du_L-1}
\begin{aligned}
\frac{\partial\mathcal{L}}{\partial \boldsymbol{u}^{l}[t]} 
= & \begin{aligned} \frac{\partial\mathcal{L}}{\partial \boldsymbol{u}^{l+1}[t]} \frac{\partial \boldsymbol{u}^{l+1}[t]}{\partial \boldsymbol{s}^{l}[t]} \frac{\partial \boldsymbol{s}^{l}[t]}{\partial \boldsymbol{u}^{l}[t]}\end{aligned}  \\
&+\sum_{t^{\prime}=t+1}^T \frac{\partial\mathcal{L}}{\partial \boldsymbol{u}^{l+1}[t^{\prime}]} \frac{\partial \boldsymbol{u}^{l+1}[t^{\prime}]}{\partial \boldsymbol{s}^{l}[t^{\prime}]} \frac{\partial \boldsymbol{s}^{l}[t^{\prime}]} {\partial \boldsymbol{u}^{l}[t^{\prime}]}\prod_{t^{\prime\prime}=1}^{t^{\prime}-t}\epsilon^{l}[t^{\prime}-t^{\prime\prime}].
\end{aligned}
\end{equation}

Finally, combining Eq.\eqref{eq:appdix:1<=dL_dU_t<=T}-\eqref{eq:appendix:dL_du_L-1}, we conclude the following equations:
\begin{equation}
\frac{\partial\mathcal{L}}{\partial \boldsymbol{u}^{l}[t]}  = \frac{\partial\mathcal{L}}{\partial \boldsymbol{s}^{l}[t]}  \frac{\partial \boldsymbol{s}^{l}[t]}{\partial \boldsymbol{u}^{l}[t]}
+ \sum_{t^{\prime}=t+1}^{T}
\frac{\partial\mathcal{L}}{\partial \boldsymbol{s}^{l}[t^{\prime}]} \frac{\partial \boldsymbol{s}^{l}[t^{\prime]}]}{\partial \boldsymbol{u}^{l}[t^{\prime]}]}
\prod_{t^{\prime\prime}=1}^{t^{\prime}-t} \boldsymbol{\epsilon}^{l}[t^{\prime}-t^{\prime\prime}]  ,
\label{eq:appendx:dL_dU_final}
\end{equation}
where
\begin{equation}\label{eq:appendx:dL_dS_final}
\frac{\partial\mathcal{L}}{\partial \boldsymbol{s}^{l}[t]} = 
\begin{cases} 
\frac{\partial\mathcal{L}}{\partial \boldsymbol{s}^{L}[t]}  & \text{ if } l = \mathrm{L} \\ 
\frac{\partial\mathcal{L}}{\partial \boldsymbol{u}^{l+1}[t]}\frac{\partial{\boldsymbol{u}^{l+1}[t]}}{\partial \boldsymbol{s}^{l}[t]}  & \text{ if } l = \mathrm{L}-1,\cdots, 1
\end{cases},
\end{equation}
The Eq.\eqref{eq:appendx:dL_dU_final} and Eq.\eqref{eq:appendx:dL_dS_final} is the same as Eq.\eqref{eq:preliminary:dL_dU} and Eq.\eqref{eq:preliminary:dL_dS}.
\end{proof}

\newpage
\section{The Details of Experimental Observation}\label{apx: The detail of Experimental Observation}

\begin{figure}[htb]
\centering
\subfigure[CIFAR10 / ResNet18]{\includegraphics[
width=.38 \textwidth, 
]
{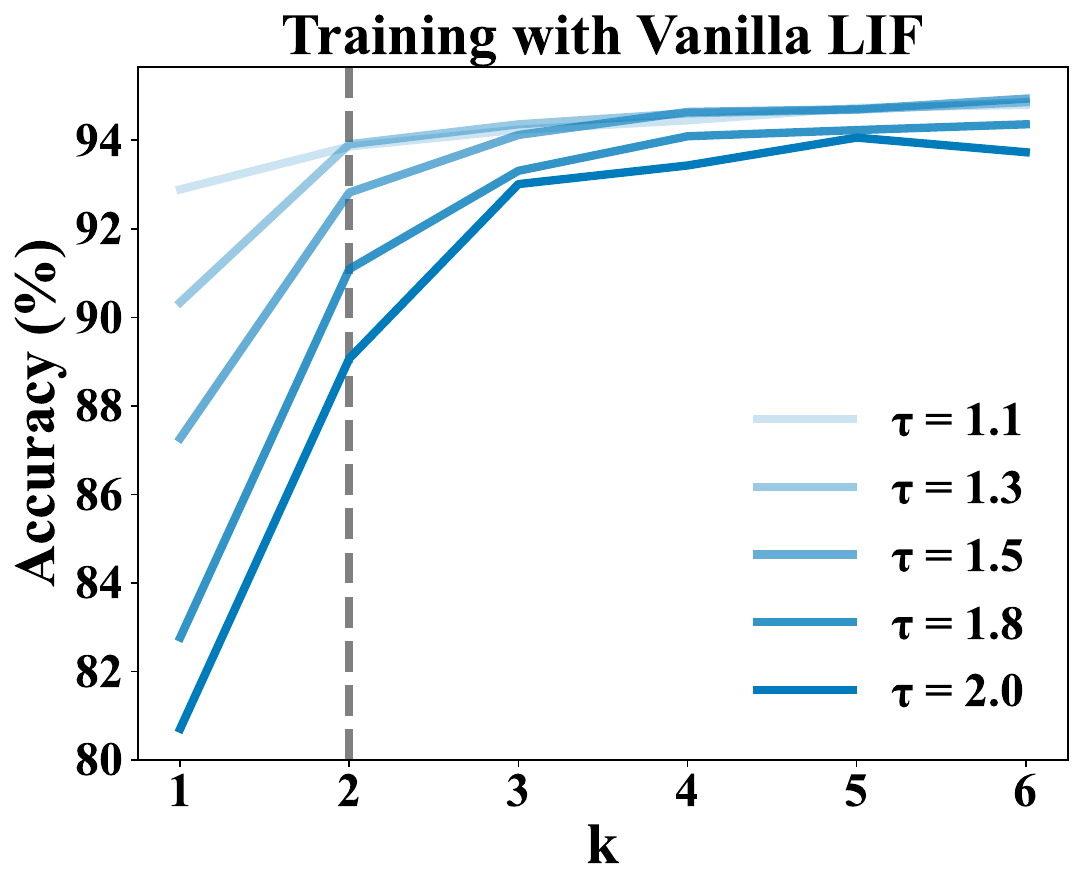}
} 
\subfigure[DVSGesture / Vgg11] {\includegraphics[
width=.38 \textwidth, 
]
{figure/appendix_img/dvsgesture_spiking_vgg11_bn_T20_K_Tau_2D.pdf}
} 
\caption{Performance of LIF with BPTT training, varying $\tau$ and $k$ (detach all gradients except $0\sim k$), on (a) CIFAR10 / ResNet18 (Timestep = 6) and (b) DVS Gesture / Vgg11 (Timestep = 20). The detailed results as shown in Table.\ref{table:appendix:detial_data_experiment_observe}, and the experiment hyper-parameter as shown in Table.\ref{tab:appendix:combined}.}
 \label{fig:appendx:obervervation_tau_vs_k}
\end{figure}

The network accuracy is influenced by the time constant ($\tau$) and BPTT timestep ($k$). Besides the 5-layer NN in Figure.\ref{fig:appendx:obervervation_tau_vs_k}, we also evaluate CIFFAR10 with Spiking ResNet18 and DVSGesture with Spiking VGG11. Similar to 5layer SNN, the gradient from further timestep could not contribute to the backpropagation training process, as increasing $k$ above 3 does not substantially enhance the accuracy. The training settings are shown in Table.\ref{tab:appendix:combined}:

\begin{figure}[htb]
\centering
{\includegraphics[width=.6 \textwidth, trim=0 50 0 50, clip]{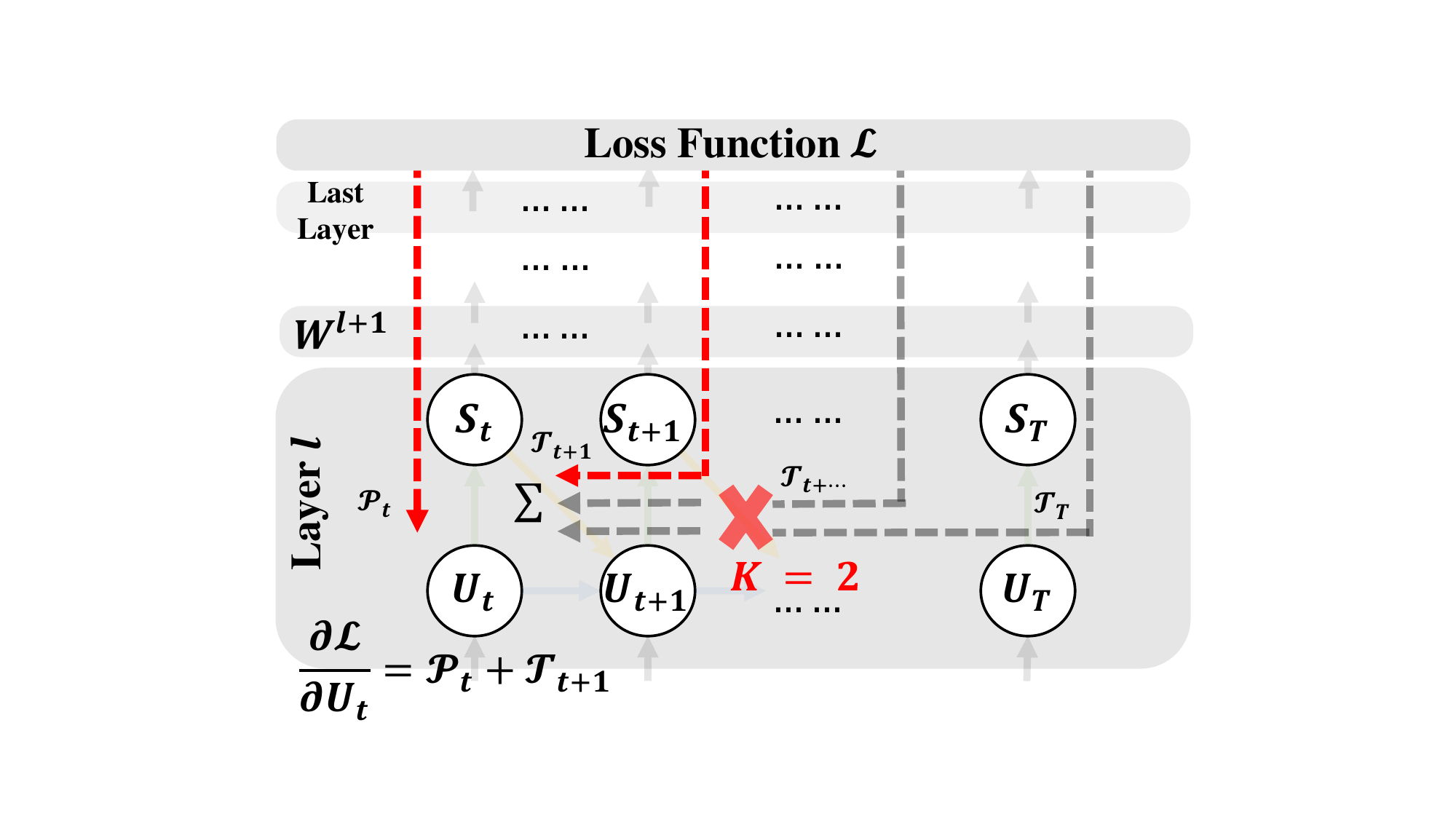}}  
\caption{Illustration of the LIF neuron based SNN’s gradient error flow during BPTT. In this example k=2: only the backpropagation from the first two timestep is considered (illustrated by the two red dashed arrows), and backpropagation along further timestep is discarded.}
 \label{fig:appendx:ob_experiment_setup}
\end{figure}

\begin{table}[ht!]\label{table:appendix:detial_data_experiment_observe}
\centering
\caption{\textbf{Left} table is the CIFAR10 accuracy performance (\%), \textbf{Right} table is the DVS Gesture accuracy performance (\%). The random seeds are uniformly fixed across all instances.}
\label{tab:appendix:combined}

\begin{tabular}{c|ccccc||c|ccccc}
\toprule
\diagbox{$k$}{$\tau$}  & 1.1   & 1.3   & 1.5   & 1.8   & 2 &  \diagbox{$k$}{$\tau$}   & 1.1   & 1.3   & 1.5   & 1.8   & 2      \\
\midrule
1 & 92.89 & 90.34 & 87.27 & 82.77 & 80.71 & 1 & 96.88 & 97.22 & 95.83 & 95.49 & 96.53  \\  \hline
2 & 93.86 & 93.91 & 92.82 & 91.09 & 89.07 & 4 & 97.57 & 97.92 & 97.22 & 96.53 & 96.53   \\ \hline
3 & 94.24 & 94.36 & 94.12 & 93.31 & 93.01 & 8  & 97.57 & 97.57 & 97.22 & 97.57 & 97.22  \\ \hline
4 & 94.45 & 94.61 & 94.64 & 94.09 & 93.43 & 12 & 97.22 & 97.57 & 97.22 & 97.22 & 97.57  \\ \hline
5 & 94.73 & 94.69 & 94.7  & 94.23 & 94.06 & 16 & 97.57 & 97.57 & 97.92 & 97.22 & 97.57  \\ \hline
6 & 94.80  & 94.86 & 94.94 & 94.36 & 93.73 & 20 & 97.57 & 97.57 & 97.92 & 97.57 & 97.22   \\
\bottomrule
\end{tabular}
\end{table}

\begin{table}[ht]
\centering
\caption{Training Parameters}
\label{tab:appendix:combined}
\begin{tabular}{l|cc}
\toprule
Parameter \ Datasets & CIFAR10 & DVS Gesture \\
\midrule
Networks & Spiking ResNet18   &  Spiking Vgg11    \\
Time Steps (T) & 6 & 20 \\
Epochs (e) & 200 & 300 \\
Batch Size (bs) & 128 & 16 \\
Optimizer & SGD & SGD \\
Learning Rate (lr) & 0.1 & 0.1 \\
Weight Decay (wd) & $5\times10^{-5}$ & $5\times10^{-4}$ \\
Dropout Rate & 0.0 & 0.4 \\
\bottomrule
\end{tabular}
\end{table}

\newpage

\section{Detailed Discussion on Temporal Gradient Errors}
\label{apx:Detailed Discussion on Temporal Gradient Errors}

Eq.\eqref{eq:preliminary:epsilon_after} can be substituted into formula Eq.\eqref{eq:observe:temporal gradients errors} yields:
\begin{equation}
\mathcal{T}^l[t, t^{\prime}] 
= \frac{\partial\mathcal{L}}{\partial \boldsymbol{s}^{l}[t^{\prime}]} 
\frac{\partial \boldsymbol{s}^{l}[t^{\prime}]}{\partial \boldsymbol{u}^{l}[t^{\prime}]}
\underbrace{\gamma^{(t^{\prime}-t)} }_{\text{Part I}}
\prod_{t^{\prime\prime}=1}^{t^{\prime}-t} \underbrace{\left( 1 - V_{\mathrm{th}} \mathbb{H}\left(\boldsymbol{u}^{l}[t^{\prime}-t^{\prime\prime}] \right) \right)  }_{\text{Part II}} ,
\end{equation}

In this case, the $\boldsymbol \epsilon$ still converges to 0 after continuous multiplication. The detailed proof is shown below.

We construct the function 
$f_{\epsilon}(n) = \prod_{t=1}^{n} \boldsymbol{\epsilon}^{l}[t], n=1, 2, \dots \label{eq:appendix:f_epsilon_n} 
$, to proof $ \lim_{n \to +\infty} f_{\epsilon}(n) = 0 $.

\begin{proof}
\begin{equation}
\begin{aligned}
\lim_{n \to +\infty}f_{\epsilon}(n) 
&= \lim_{n \to +\infty} \prod_{t=1}^{n} \boldsymbol{\epsilon}^{l}[t] \\
&= \lim_{n \to +\infty} \prod_{t=1}^{n} \gamma \left( 1 -  \boldsymbol{s}^l[t] \odot \mathbb{H} \left( \boldsymbol{u}^l[t] \right) \right) \\
&=  \lim_{n \to +\infty} \gamma^{n} \prod_{t=1}^{n}\left( 1 -  \boldsymbol{s}^l[t] \odot \mathbb{H} \left( \boldsymbol{u}^l[t] \right) \right) \\
&= 0 
\end{aligned}
\end{equation}
where the $\gamma$ is a constant and $0 < \gamma < 1$, resulting $ \lim_{n \to +\infty} \gamma^{n} = 0$.
\end{proof}

\begin{equation}\label{eq:appendix:partII}
1- V_{\mathrm{th}} \mathbb{H}(\mathbf{u}^l\ [t]) = 
\begin{cases}
1- \frac{V_{\mathrm{th}}}{\alpha},& \text{if } V_{\mathrm{th}} - \frac{\alpha}{2} \leq  u^l [t] \leq V_{\mathrm{th}} + \frac{\alpha}{2} \\ 1, & \text{otherwise}
\end{cases}.
\end{equation}
In Part II, we further deduce as shown in Eq.\eqref{eq:appendix:partII}, drawing from Eq.\eqref{eq:preliminary:eligibility}. The second term also tends toward zero, influenced by the hyperparameter $\alpha$. To prevent spatial gradient explosion, $\alpha$ is typically set to be larger than or equal to  $V_{\mathrm{th}}$ (Wu et al., 2019). When $\alpha > V_{\mathrm{th}}$, the result is $0<1- \frac{V_{\mathrm{th}}}{\alpha} < 1$, which causes the temporal gradients to converge to zero more quickly due to the continuous product. However, many studies retain the default value of $\alpha = V_{\mathrm{th}} = 1$ \cite{deng2021temporal}. When $\alpha = V_{\mathrm{th}}$,  if the membrane potential is within the range of $ (V_{\mathrm{th}} - \frac{\alpha}{2},  V_{\mathrm{th}} + \frac{\alpha}{2})$, then $1 - \frac{V_{\mathrm{th}}}{\alpha}$ equals. In other words, if a spike is generated (or $u \approx V_{\mathrm{th}}$) once within the range of $(t+1, T)$, the temporal gradient will be 0 in such cases. This signifies a pervasive challenge with temporal gradients in the vanilla LIF model, persisting even with short timestep.

{\color{black}
\section{Detailed Discussion on Inspired Biological Principles} \label{apx:Detailed Discussion on inspired biological principles}

The design inspiration for CLIF neurons primarily comes from the adaptive learning characteristics observed in the biological nervous system, particularly the mechanisms of neural adaptability and dynamic regulation of membrane potential \cite{benda2003universal}. In biology, neurons adjust their electrophysiological properties to adapt to different environmental stimuli. This capability is crucial for the effective processing of information by neurons, preventing excessive excitability \cite{klausberger2008neuronal, rankin2009habituation}. A key biological mechanism is regulation through the activity of GABAergic neurons, which release GABA onto the postsynaptic membrane of the target neuron, leading to hyperpolarization and inhibition of excessive action potential production \cite{letinic2002origin}.

The CLIF model simulates this hyperpolarization process and the regulation of action potential generation by resetting a greater amount of membrane potential after each firing, attempting to replicate this type of adaptive regulation characteristic of biological neurons in a computational model.
}

\section{Dynamic Analysis of CLIF neuron} \label{apx: The dynamic of CLIF neuron}
In this section, we first discuss the firing dynamic behavior of the CLIF, and then we discuss the auto-correlation for the membrane potential. Finally, we discuss the dynamic difference between the CLIF and the current-based, adaptive threshold model.

\begin{figure}[htbp]
	\centering
	\subfigure[Timestep = 20] {\includegraphics[width=.40\textwidth]{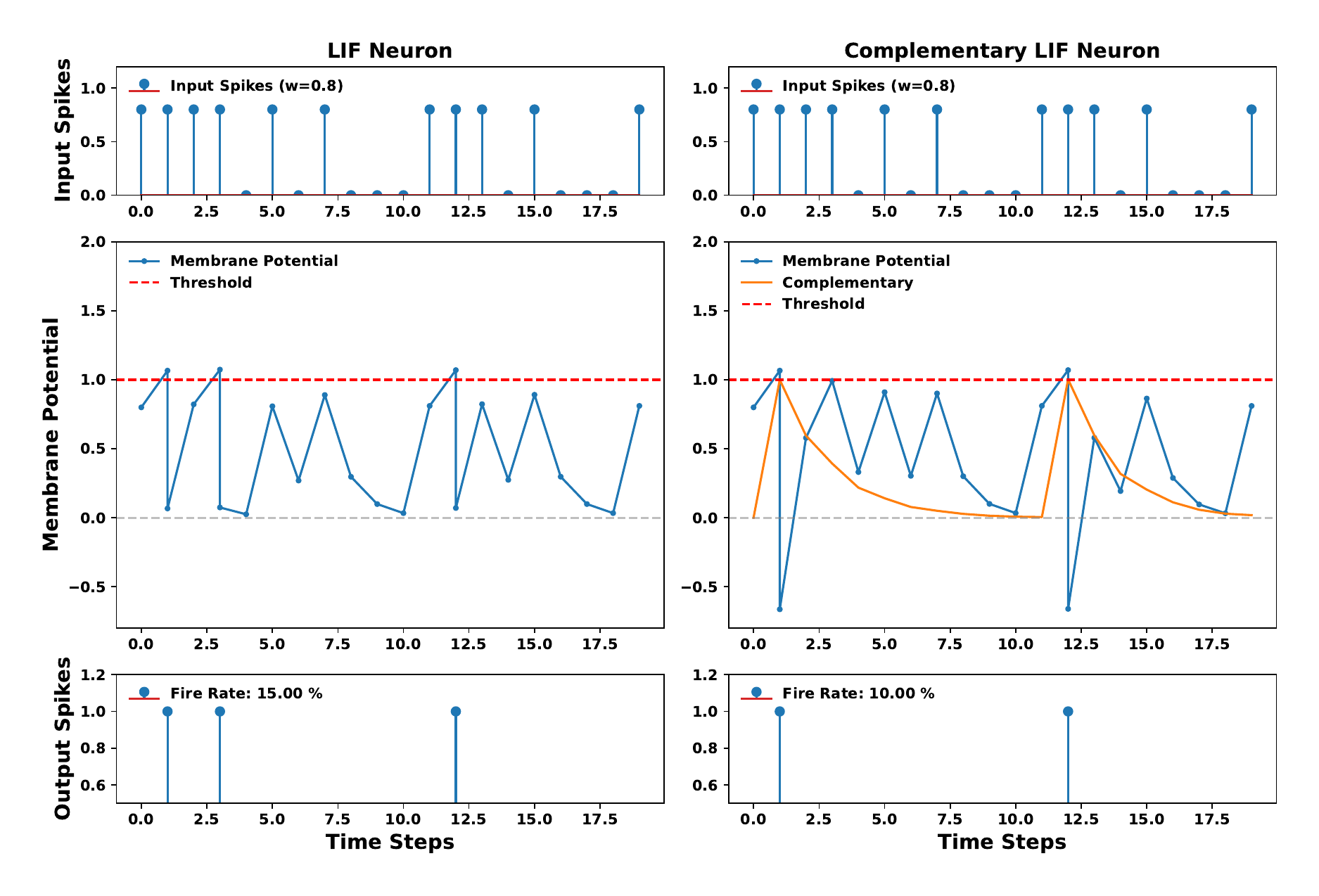}}
	\subfigure[Timestep = 40] {\includegraphics[width=.40\textwidth]{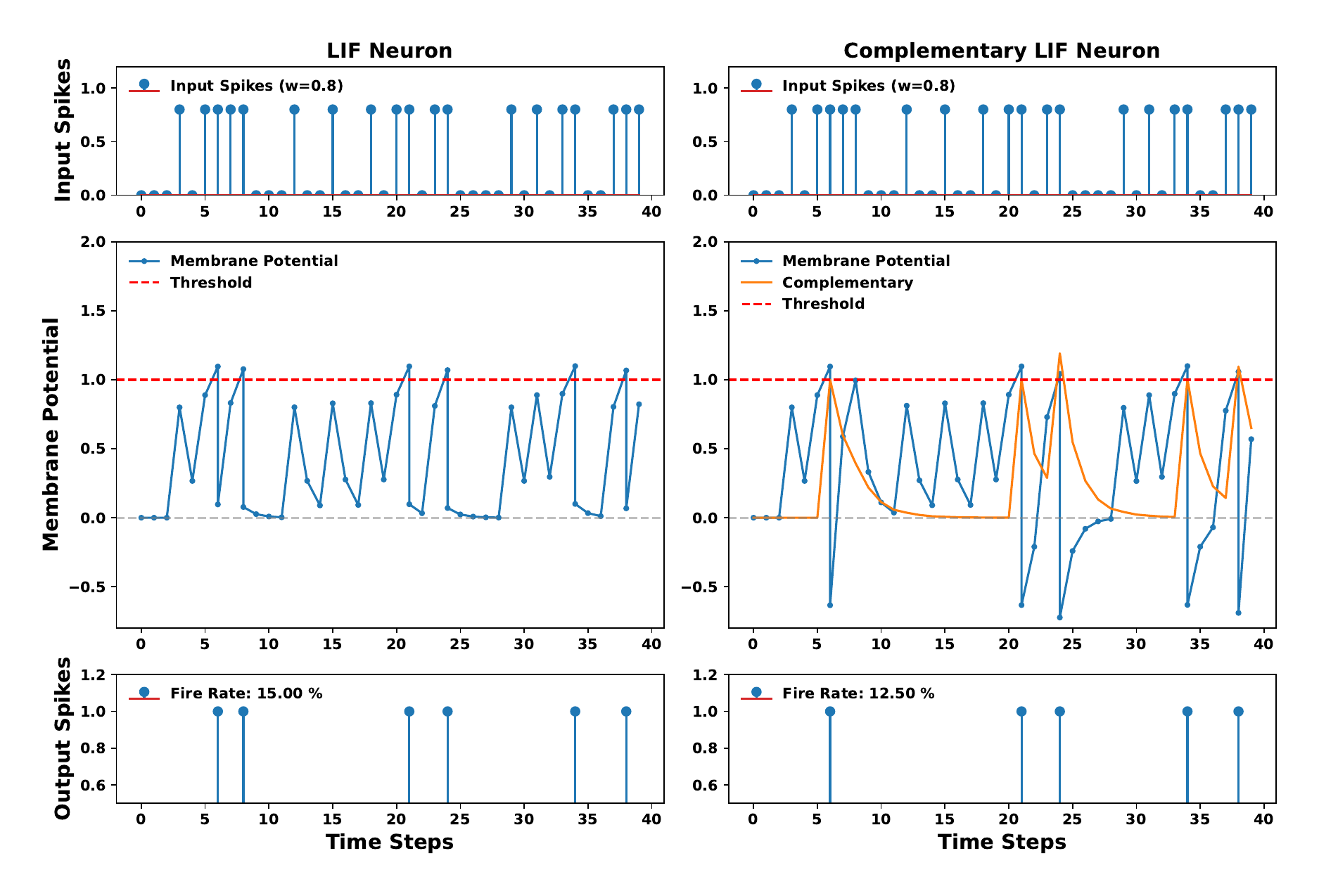}}
	\caption{The dynamic behavior of a single LIF and CLIF neuron at different timestep.}
	\label{figure:appdix:Dynamic Analysis of CLIF neuron}
\end{figure}

Firstly, we analyze the fire rate and auto-correlation of CLIF according to the same Poisson random input. The firing dynamic behavior under the different timestep for the single CLIF neuron is shown in Figure.\ref{figure:appdix:Dynamic Analysis of CLIF neuron}. We can find that compared with LIF neurons, CLIF neuron has extra refractory periods resulting lower fire rate. 

{\color{black} Secondly, we calculate the auto-correlation function using $R_x[k] = \frac{1}{N} \sum_{n=0}^{N-1} x[n] \cdot x[n-k]$. The results are shown in Figure.\ref{figure:appdix:Auto-correlation of CLIF neuron}. The auto-correlation function indicates the degree of correlation between a signal and a delayed version of itself. We can observe that the auto-correlation value of the complementary membrane potential decays slower, and its period is longer. This suggests that CLIF can capture more and longer correlations in the temporal dimension than LIF.

\begin{figure}[htbp]
	\centering
	\subfigure[Timestep = 20] {\includegraphics[width=.40\textwidth]{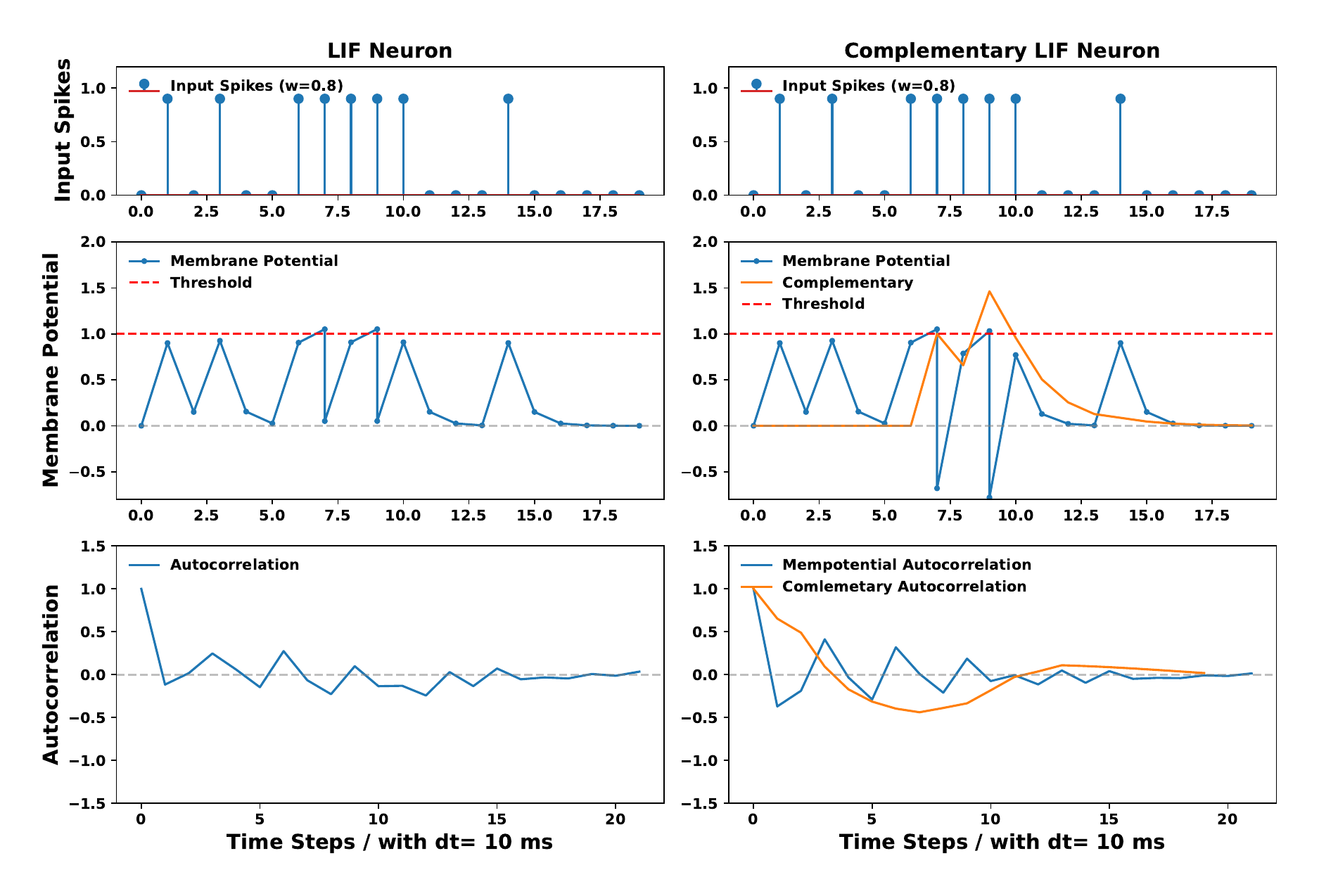}}
	\subfigure[Timestep = 40] {\includegraphics[width=.40\textwidth]{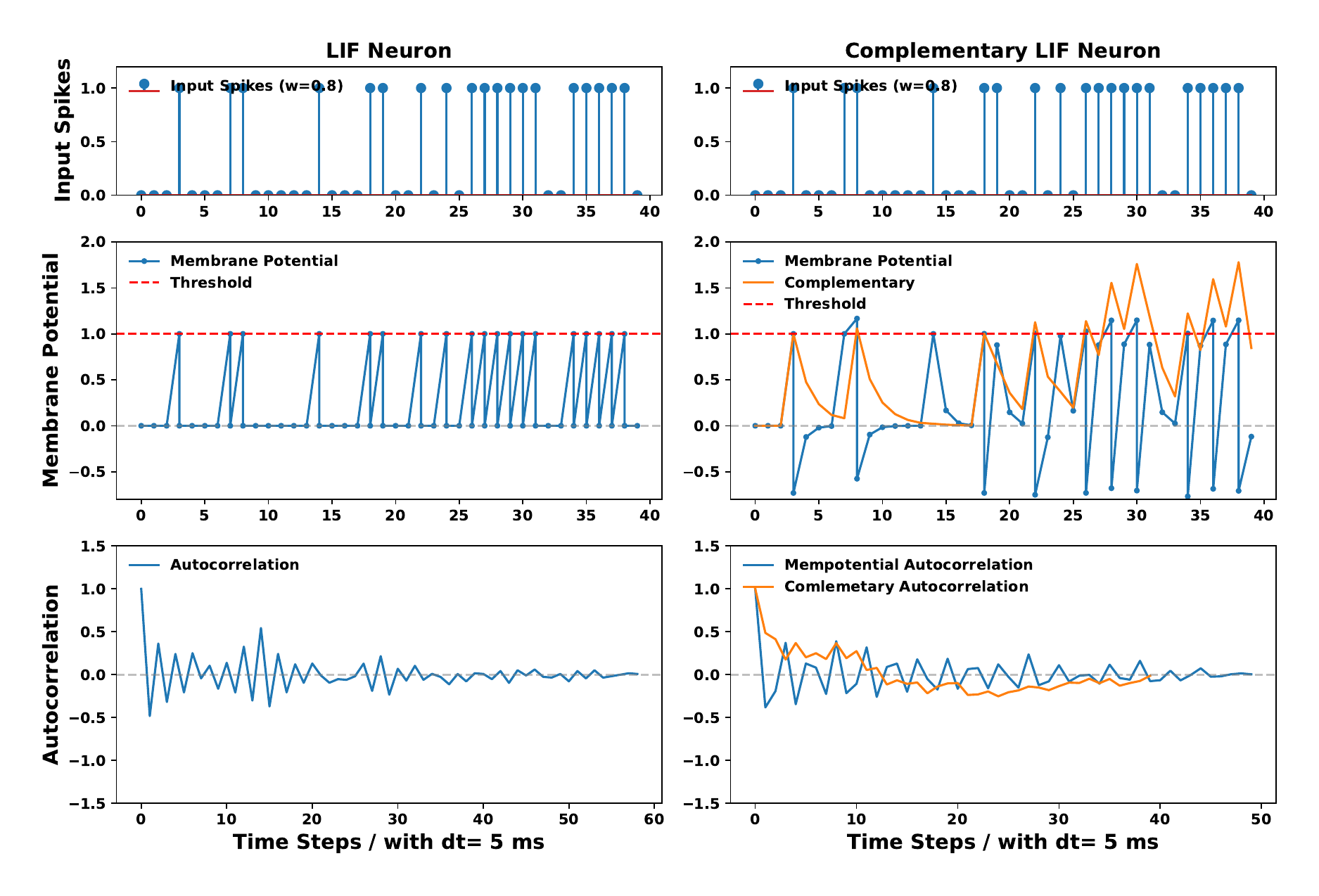}}
	\caption{The autocorrelation of a single LIF and CLIF neuron at different timestep.}
	\label{figure:appdix:Auto-correlation of CLIF neuron}
\end{figure}

Finally, CLIF shares more similarities to the Adaptive Threshold model \cite{bellec2020solution} than to the Current-Base model \cite{zenke2018superspike}. As for synaptic input current for both CLIF and adaptive threshold model take the form of $ \boldsymbol{W} \boldsymbol{s}[t] $, different from current-base model $ \boldsymbol{I}_{\mathrm{syn}}[t] = \frac{1}{\tau}\boldsymbol{I}_{\mathrm{syn}}[t-1] + \boldsymbol{W} \boldsymbol{s}[t] $, adaptive threshold model uses a latent variable to adjust the neurons' firing thresholds, whereas CLIF uses a latent variable (the complementary membrane potential) to adjust neurons reset strength.
}

\section{The Gradients of CLIF Neuron} 
\label{apx:The gradients of CLIF model}
The CLIF model can be rewritten as:
\begin{gather}
\boldsymbol{u}^l[t] = 
\gamma \left( \boldsymbol{u}^l[t-1] - \boldsymbol{s}^l[t-1] \odot \left(V_{\mathrm{th}} + \sigma(\boldsymbol{m}^l[t-1]) \right) \right) + \boldsymbol{W}^{l}\boldsymbol{s}^{l-1}[t]\\
\boldsymbol{s}^l[t] = \Theta(\boldsymbol{u}^l[t] - V_{\mathrm{th}})  \\
\boldsymbol{m}^l[t] = \boldsymbol{m}^l[t-1] \odot \sigma\left(\frac{1}{\tau} \boldsymbol{u}^l[t]\right) + \boldsymbol{s}^l[t]    
\end{gather}
where defined $\gamma \equiv 1-\frac{1}{\tau}$, then the gradients at $l$ layer is calculated as:
\begin{equation}
\nabla_{\boldsymbol{W}^l}\mathcal{L}=\sum_{t=1}^T \frac{\partial\mathcal{L}}{\partial \boldsymbol{u}^{l}[t]} \frac{\partial \boldsymbol{u}^{l}[t]}{\partial \boldsymbol{W}^{l}} ,l=\mathrm{L},\mathrm{L}-1,\cdots,1.
\end{equation}
Where the right part could be deduced as:
\begin{equation}
\frac{\partial \boldsymbol{u}^{l}[t]}{\partial \boldsymbol{W}^{l}} =  \boldsymbol{s}^{l-1}[t]  
\end{equation}

For the left part, we recursively evaluate:
\begin{equation}
\label{aeq:dL_du}
\frac{\partial\mathcal{L}}{\partial \boldsymbol{u}^{l}[t]} 
= \underbrace{\frac{\partial\mathcal{L}}{\partial \boldsymbol{s}^{l}[t]} \frac{\partial  \boldsymbol{s}^{l}[t]}{\partial \boldsymbol{u}^{l}[t]}}_{\text{Spatial Gradients}}
+ 
\underbrace{
\frac{\partial\mathcal{L}}{\partial \boldsymbol{u}^{l}[t+1]} 
\left( 
\boldsymbol{\epsilon}^{l}[t] + 
\frac{\partial \boldsymbol{u}^{l}[t+1]}{\partial \boldsymbol{m}^{l}[t]} 
\boldsymbol{\psi}^{l}[t] \right)}_{ \text{Temporal Gradients of Membrane Potential}} 
+ 
\underbrace{\frac{\partial\mathcal{L}}{\partial \boldsymbol{m}^{l}[t]} 
\boldsymbol{\psi}^{l}[t] }_{\text{Temporal Gradients of Complementary}}
\end{equation}

The eligibility, this terminology mainly refers to e-prop \cite{bellec2020solution}, STBP \cite{wu2018spatio}. From the LIF model, the equation can be deduced as:
\begin{equation}
\boldsymbol{\epsilon}^{l}[t] \equiv \frac{\partial \boldsymbol{u}^{l}[t+1]}{\partial \boldsymbol{u}^{l}[t]} + \frac{\partial \boldsymbol{u}^{l}[t+1]}{\partial \boldsymbol{s}^{l}[t]} \frac{\partial \boldsymbol{s}^{l}[t]}{\partial \boldsymbol{u}^{l}[t]}
\end{equation}

The Complementary will also introduce the $\boldsymbol{\psi}^{l}[t]$:
\begin{equation}
\begin{aligned}
\boldsymbol{\psi}^{l}[t] 
\equiv  &
\frac{\partial \boldsymbol{m}^{l}[t]}{\partial \boldsymbol{u}^{l}[t]} +
\frac{\partial \boldsymbol{m}^{l}[t]}{\partial \boldsymbol{s}^{l}[t]} 
\frac{\partial \boldsymbol{s}^{l}[t]}{\partial \boldsymbol{u}^{l}[t]}  \\
= &
\underbrace{\frac{1}{\tau} \boldsymbol{m}^l[t-1]}_{\geq 0} \odot \underbrace{\sigma^{\prime}\left(\frac{1}{\tau} \boldsymbol{u}^l[t]\right)}_{\in (0, 1)} + \underbrace{\mathbb{H}\left(\boldsymbol{u}^{l}[t]\right)}_{\geq 0} 
\end{aligned}
\end{equation}


Besides, the Complementary gradient line will introduce the recursively part:
\begin{equation}
\label{aeq:dL_dm}
\frac{\partial\mathcal{L}}{\partial \boldsymbol{m}^{l}[t]} 
= 
\frac{\partial\mathcal{L}}{\partial \boldsymbol{u}^{l}[t+1]} 
\frac{\partial \boldsymbol{u}^{l}[t+1]}{\partial \boldsymbol{m}^{l}[t]}
+
\frac{\partial\mathcal{L}}{\partial \boldsymbol{m}^{l}[t+1]} 
\left(
\frac{\partial \boldsymbol{m}^{l}[t+1]}{\partial \boldsymbol{m}^{l}[t]}
+
\boldsymbol{\psi}^{l}[t+1] 
\frac{\partial \boldsymbol{u}^{l}[t+1]}{\partial \boldsymbol{m}^{l}[t]}
\right)
\end{equation}

\newpage 

To better understand the eligibility in Eq.\ref{aeq:dL_du} and Eq.\ref{aeq:dL_dm}, we can refer to the following Figure.\ref{fig:appendix:forward_dependency_backward_eligibility}:
\begin{figure*}[hb]
\centering
\includegraphics[width=0.68\textwidth,trim=0 90 0 90,clip]{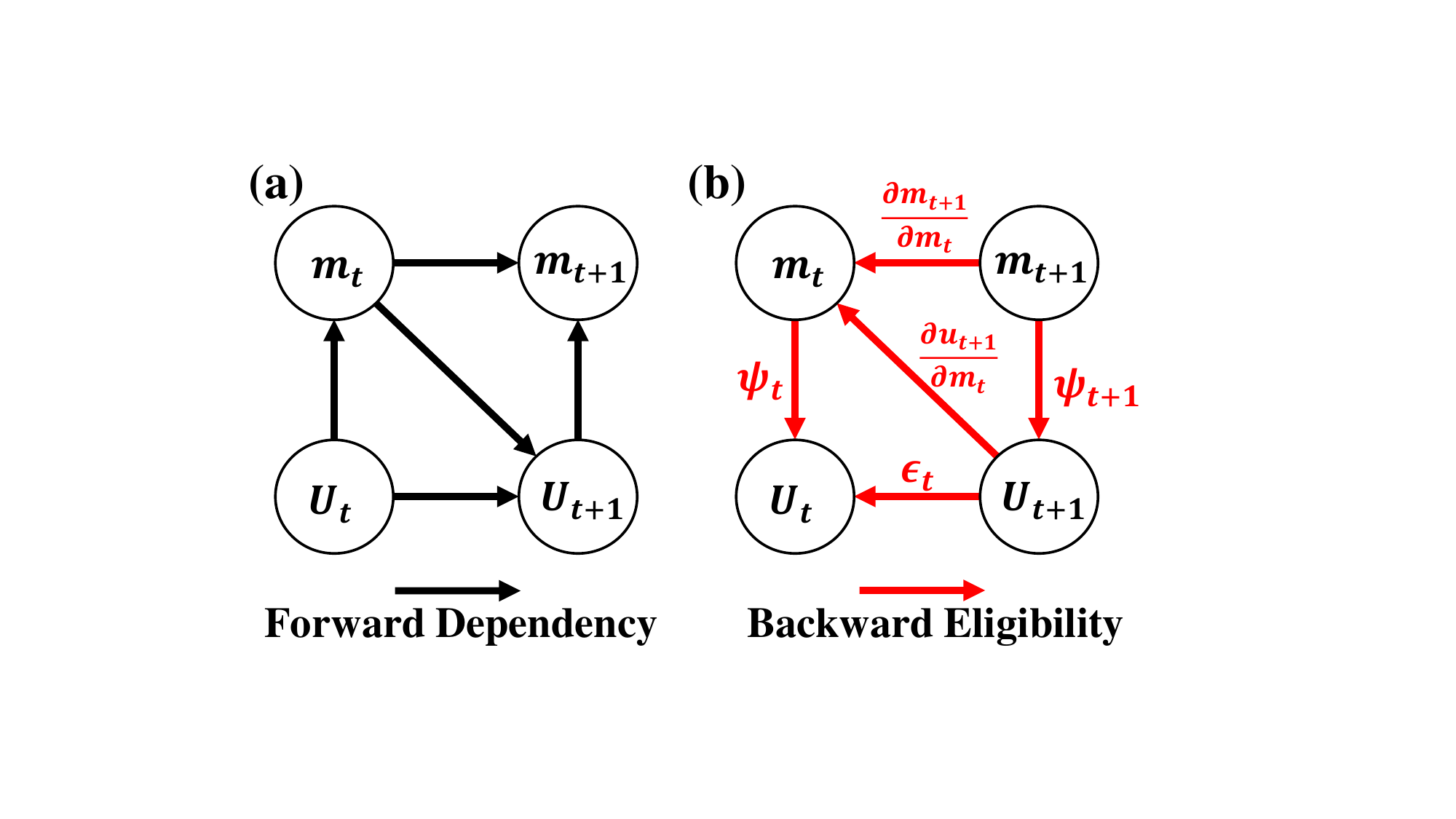} 
\caption{The abstract expression of (a) forward dependency and (b) backward eligibility trace for CLIF neuron.}
\label{fig:appendix:forward_dependency_backward_eligibility}
\end{figure*}

Merging the Eq.\ref{aeq:dL_du} and Eq.\ref{aeq:dL_dm} as matrix computing process:
\begin{equation}
\begin{bmatrix}
{\color{blue} 1} & -\boldsymbol{\psi}^{l}[t] \\
0 & 1
\end{bmatrix}
\begin{bmatrix}
{\color{blue}\frac{\partial\mathcal{L}}{\partial \boldsymbol{u}^{l}[t]}} \\
\frac{\partial\mathcal{L}}{\partial \boldsymbol{m}^{l}[t]}
\end{bmatrix}
=
\begin{bmatrix}
{\color{blue}\boldsymbol{\epsilon}^{l}[t]} +  \frac{\partial \boldsymbol{u}^{l}[t+1]}{\partial \boldsymbol{m}^{l}[t]} \boldsymbol{\psi}^{l}[t]    &  0\\
\frac{\partial \boldsymbol{u}^{l}[t+1]}{\partial \boldsymbol{m}^{l}[t]}  
& \frac{\partial \boldsymbol{m}^{l}[t+1]}{\partial \boldsymbol{m}^{l}[t]}
+
\boldsymbol{\psi}^{l}[t+1] 
\frac{\partial \boldsymbol{u}^{l}[t+1]}{\partial \boldsymbol{m}^{l}[t]}
\end{bmatrix}
\begin{bmatrix}
{\color{blue}\frac{\partial\mathcal{L}}{\partial \boldsymbol{u}^{l}[t+1]}} \\
\frac{\partial\mathcal{L}}{\partial \boldsymbol{m}^{l}[t+1]}
\end{bmatrix}
+
\begin{bmatrix}
{\color{blue}\frac{\partial\mathcal{L}}{\partial \boldsymbol{s}^{l}[t]} \frac{\partial  \boldsymbol{s}^{l}[t]}{\partial \boldsymbol{u}^{l}[t]}} \\
0
\end{bmatrix}
\end{equation}
where {\color{blue}the blue part is the original gradient of the LIF neuron}, and the other parts are introduced by Complementary.

Firstly, we first recursively expand the Eq.\ref{aeq:dL_dm}:
\begin{equation}
\begin{aligned}
\frac{\partial\mathcal{L}}{\partial \boldsymbol{m}^{l}[t]} 
=& 
\frac{\partial\mathcal{L}}{\partial \boldsymbol{u}^{l}[t+1]} 
\frac{\partial \boldsymbol{u}^{l}[t+1]}{\partial \boldsymbol{m}^{l}[t]}
+
\frac{\partial\mathcal{L}}{\partial \boldsymbol{m}^{l}[t+1]} 
\underbrace{
\left(
\frac{\partial \boldsymbol{m}^{l}[t+1]}{\partial \boldsymbol{m}^{l}[t]}
+
\boldsymbol{\psi}^{l}[t+1] 
\frac{\partial \boldsymbol{u}^{l}[t+1]}{\partial \boldsymbol{m}^{l}[t]}
\right) }_{\text{Define as } \boldsymbol{\rho}^{l}[t]}  \\
=&
\frac{\partial\mathcal{L}}{\partial \boldsymbol{u}^{l}[t+1]} 
\frac{\partial \boldsymbol{u}^{l}[t+1]}{\partial \boldsymbol{m}^{l}[t]}
+
\underbrace{ \left(
\frac{\partial\mathcal{L}}{\partial \boldsymbol{u}^{l}[t+2]} 
\frac{\partial \boldsymbol{u}^{l}[t+2]}{\partial \boldsymbol{m}^{l}[t+1]}
+
\frac{\partial\mathcal{L}}{\partial \boldsymbol{m}^{l}[t+2]} 
\boldsymbol{\rho}^{l}[t+1]
\right) }_{\text{expansion}}
\boldsymbol{\rho}^{l}[t]  \\
=&
\frac{\partial\mathcal{L}}{\partial \boldsymbol{u}^{l}[t+1]} 
\frac{\partial \boldsymbol{u}^{l}[t+1]}{\partial \boldsymbol{m}^{l}[t]}
+
\frac{\partial\mathcal{L}}{\partial \boldsymbol{u}^{l}[t+2]} 
\frac{\partial \boldsymbol{u}^{l}[t+2]}{\partial \boldsymbol{m}^{l}[t+1]} \boldsymbol{\rho}^{l}[t]
+
\underbrace{
\frac{\partial\mathcal{L}}{\partial \boldsymbol{m}^{l}[t+2]} }_{\text{expansion}}
\boldsymbol{\rho}^{l}[t+1]
\boldsymbol{\rho}^{l}[t]  \\
=&
\frac{\partial\mathcal{L}}{\partial \boldsymbol{u}^{l}[t+1]} 
\frac{\partial \boldsymbol{u}^{l}[t+1]}{\partial \boldsymbol{m}^{l}[t]}
+
\frac{\partial\mathcal{L}}{\partial \boldsymbol{u}^{l}[t+2]} 
\frac{\partial \boldsymbol{u}^{l}[t+2]}{\partial \boldsymbol{m}^{l}[t+1]} \boldsymbol{\rho}^{l}[t] +
\frac{\partial\mathcal{L}}{\partial \boldsymbol{u}^{l}[t+3]} 
\frac{\partial \boldsymbol{u}^{l}[t+3]}{\partial \boldsymbol{m}^{l}[t+2]} 
\boldsymbol{\rho}^{l}[t+1] 
\boldsymbol{\rho}^{l}[t] \\
&+
\underbrace{
\frac{\partial\mathcal{L}}{\partial \boldsymbol{m}^{l}[t+3]} }_{\text{expansion until T}}
\boldsymbol{\rho}^{l}[t+2]
\dots
\boldsymbol{\rho}^{l}[t]  \\
=&
\frac{\partial\mathcal{L}}{\partial \boldsymbol{u}^{l}[t+1]} 
\frac{\partial \boldsymbol{u}^{l}[t+1]}{\partial \boldsymbol{m}^{l}[t]} 
+
\frac{\partial\mathcal{L}}{\partial \boldsymbol{u}^{l}[t+2]} 
\frac{\partial \boldsymbol{u}^{l}[t+2]}{\partial \boldsymbol{m}^{l}[t+1]} \boldsymbol{\rho}^{l}[t] 
+ ...
+
\frac{\partial\mathcal{L}}{\partial \boldsymbol{u}^{l}[T]} 
\frac{\partial \boldsymbol{u}^{l}[T]}{\partial \boldsymbol{m}^{l}[T-1]} \boldsymbol{\rho}^{l}[T-2] ... \boldsymbol{\rho}^{l}[t+1] \boldsymbol{\rho}^{l}[t]  \\
\end{aligned}
\end{equation}

By mathematical induction, we can deduce that:
\begin{equation}\label{aeq:dL_dm_final_expression}
\begin{aligned}
\frac{\partial\mathcal{L}}{\partial \boldsymbol{m}^{l}[t]} 
=& 
\sum_{t^{\prime}=t+1}^{T} 
\frac{\partial\mathcal{L}}{\partial \boldsymbol{u}^{l}[t^{\prime}]} 
\frac{\partial \boldsymbol{u}^{l}[t^{\prime}]}{\partial \boldsymbol{m}^{l}[t^{\prime}-1]} 
\prod_{t^{\prime\prime}=2}^{t^{\prime}-t} \boldsymbol{\rho}^{l}[t^{\prime} - t^{\prime\prime}]  \\
\end{aligned}
\end{equation}

Secondly, we recursively expand the Eq.\eqref{aeq:dL_du}:
\begin{equation}\label{aeq:dL_du_recursively}
\begin{aligned}
\frac{\partial\mathcal{L}}{\partial \boldsymbol{u}^{l}[t]} 
=& 
\underbrace{\frac{\partial\mathcal{L}}{\partial \boldsymbol{s}^{l}[t]} 
\frac{\partial  \boldsymbol{s}^{l}[t]}{\partial \boldsymbol{u}^{l}[t]}}_{\text{Spatial Gradients}}
+ 
\underbrace{\frac{\partial\mathcal{L}}{\partial \boldsymbol{u}^{l}[t+1]} 
\left( \boldsymbol{\epsilon}^{l}[t] +  
\frac{\partial \boldsymbol{u}^{l}[t+1]}{\partial \boldsymbol{m}^{l}[t]} \boldsymbol{\psi}^{l}[t] \right)}_{ \text{Temporal Gradients of Membrane Potential}} 
+ \underbrace{\frac{\partial\mathcal{L}}{\partial \boldsymbol{m}^{l}[t]} \boldsymbol{\psi}^{l}[t] }_{\text{Temporal Gradients of Complementary}} \\
=&  
\frac{\partial\mathcal{L}}{\partial \boldsymbol{s}^{l}[t]} 
\frac{\partial  \boldsymbol{s}^{l}[t]}{\partial \boldsymbol{u}^{l}[t]}
+ 
\underbrace{\frac{\partial\mathcal{L}}{\partial \boldsymbol{u}^{l}[t+1]}}_{
\text{expansion}
}
\underbrace{\left( \boldsymbol{\epsilon}^{l}[t] +  
\frac{\partial \boldsymbol{u}^{l}[t+1]}{\partial \boldsymbol{m}^{l}[t]} \boldsymbol{\psi}^{l}[t] \right)}_{ \text{Define as $\boldsymbol{\xi}^l[t]$}} 
+ \frac{\partial\mathcal{L}}{\partial \boldsymbol{m}^{l}[t]} \boldsymbol{\psi}^{l}[t] \\
=&  
\frac{\partial\mathcal{L}}{\partial \boldsymbol{s}^{l}[t]} 
\frac{\partial  \boldsymbol{s}^{l}[t]}{\partial \boldsymbol{u}^{l}[t]}
+ 
\left(
\frac{\partial\mathcal{L}}{\partial \boldsymbol{s}^{l}[t+1]} 
\frac{\partial  \boldsymbol{s}^{l}[t+1]}{\partial \boldsymbol{u}^{l}[t+1]}
+ 
\frac{\partial\mathcal{L}}{\partial \boldsymbol{u}^{l}[t+2]} 
\boldsymbol{\xi}^l[t+1] 
+ \frac{\partial\mathcal{L}}{\partial \boldsymbol{m}^{l}[t+1]} \boldsymbol{\psi}^{l}[t+1]
\right)
\boldsymbol{\xi}^l[t] 
+ \frac{\partial\mathcal{L}}{\partial \boldsymbol{m}^{l}[t]} \boldsymbol{\psi}^{l}[t] \\
=& 
\frac{\partial\mathcal{L}}{\partial \boldsymbol{s}^{l}[t]} 
\frac{\partial  \boldsymbol{s}^{l}[t]}{\partial \boldsymbol{u}^{l}[t]}
+ 
\frac{\partial\mathcal{L}}{\partial \boldsymbol{s}^{l}[t+1]} 
\frac{\partial  \boldsymbol{s}^{l}[t+1]}{\partial \boldsymbol{u}^{l}[t+1]} 
\boldsymbol{\xi}^l[t]
+ 
\underbrace{\frac{\partial\mathcal{L}}{\partial \boldsymbol{u}^{l}[t+2]}}_{\text{expansion utill T}}
\boldsymbol{\xi}^l[t+1] 
\boldsymbol{\xi}^l[t] \\
&+ 
\frac{\partial\mathcal{L}}{\partial \boldsymbol{m}^{l}[t+1]} \boldsymbol{\psi}^{l}[t+1]
\boldsymbol{\xi}^l[t] 
+ \frac{\partial\mathcal{L}}{\partial \boldsymbol{m}^{l}[t]} \boldsymbol{\psi}^{l}[t] \\
=& 
\frac{\partial\mathcal{L}}{\partial \boldsymbol{s}^{l}[t]} 
\frac{\partial  \boldsymbol{s}^{l}[t]}{\partial \boldsymbol{u}^{l}[t]}
+ 
\frac{\partial\mathcal{L}}{\partial \boldsymbol{s}^{l}[t+1]} 
\frac{\partial  \boldsymbol{s}^{l}[t+1]}{\partial \boldsymbol{u}^{l}[t+1]} 
\boldsymbol{\xi}^l[t]
+ 
\dots
+
\frac{\partial\mathcal{L}}{\partial \boldsymbol{s}^{l}[T]}
\frac{\partial \boldsymbol{s}^{l}[T]}{\partial \boldsymbol{u}^{l}[T]} 
\boldsymbol{\xi}^l[T-1]
\dots
\boldsymbol{\xi}^l[t]  \\
&+
\frac{\partial\mathcal{L}}{\partial \boldsymbol{m}^{l}[T]}\boldsymbol{\psi}^{l}[T]
\boldsymbol{\xi}^l[T-1] 
\dots
\boldsymbol{\xi}^l[t] \\
&+
\frac{\partial\mathcal{L}}{\partial \boldsymbol{m}^{l}[T-1]} \boldsymbol{\psi}^{l}[T-1] 
\boldsymbol{\xi}^l[T-2] 
\dots
\boldsymbol{\xi}^l[t] \\
&+
\dots
+
\frac{\partial\mathcal{L}}{\partial \boldsymbol{m}^{l}[t+1]} \boldsymbol{\psi}^{l}[t+1]
\boldsymbol{\xi}^l[t] 
+ \frac{\partial\mathcal{L}}{\partial \boldsymbol{m}^{l}[t]} \boldsymbol{\psi}^{l}[t] 
\end{aligned}
\end{equation}

Note that similar items in Eq.\eqref{aeq:dL_du_recursively} can be merged as:
\begin{equation}
\begin{aligned}
\frac{\partial\mathcal{L}}{\partial \boldsymbol{u}^{l}[t]} 
=& 
\frac{\partial\mathcal{L}}{\partial \boldsymbol{s}^{l}[t]} 
\frac{\partial  \boldsymbol{s}^{l}[t]}{\partial \boldsymbol{u}^{l}[t]}
+ 
\frac{\partial\mathcal{L}}{\partial \boldsymbol{m}^{l}[t]} \boldsymbol{\psi}^{l}[t] 
+ 
\left(
\frac{\partial\mathcal{L}}{\partial \boldsymbol{s}^{l}[t+1]} 
\frac{\partial \boldsymbol{s}^{l}[t+1]}{\partial \boldsymbol{u}^{l}[t+1]} 
+
\frac{\partial\mathcal{L}}{\partial \boldsymbol{m}^{l}[t+1]} \boldsymbol{\psi}^{l}[t+1]
\right)
\boldsymbol{\xi}^l[t] \\
&+ 
\dots
+
\left(
\frac{\partial\mathcal{L}}{\partial \boldsymbol{s}^{l}[T]}
\frac{\partial \boldsymbol{s}^{l}[T]}{\partial \boldsymbol{u}^{l}[T]} 
+
\frac{\partial\mathcal{L}}{\partial \boldsymbol{m}^{l}[T]} \boldsymbol{\psi}^{l}[T]
\right)
\boldsymbol{\xi}^l[T-1]
\boldsymbol{\xi}^l[T-2]
\dots
\boldsymbol{\xi}^l[t+1] 
\boldsymbol{\xi}^l[t] \\
=& 
{\color{blue}
\frac{\partial\mathcal{L}}{\partial \boldsymbol{s}^{l}[t]}  
\frac{\partial \boldsymbol{s}^{l}[t]}{\partial \boldsymbol{u}^{l}[t]}
}
+
\frac{\partial\mathcal{L}}{\partial \boldsymbol{m}^{l}[t]} \boldsymbol{\psi}^{l}[t] 
+ 
\sum_{t^{\prime}=t+1}^{T}
\left(
{\color{blue}
\frac{\partial\mathcal{L}}{\partial \boldsymbol{s}^{l}[t^{\prime}]} 
\frac{\partial \boldsymbol{s}^{l}[t^{\prime}]}{\partial \boldsymbol{u}^{l}[t^{\prime }]}
}
+
\frac{\partial\mathcal{L}}{\partial \boldsymbol{m}^{l}[t^{\prime}]} \boldsymbol{\psi}^{l}[t^{\prime}] 
\right)
\prod_{t^{\prime\prime}=1}^{t^{\prime}-t} \boldsymbol{\xi}^{l}[t^{\prime}-t^{\prime\prime}] \\
=& 
{\color{blue}
\frac{\partial\mathcal{L}}{\partial \boldsymbol{s}^{l}[t]}  
\frac{\partial \boldsymbol{s}^{l}[t]}{\partial \boldsymbol{u}^{l}[t]}
}
+
\frac{\partial\mathcal{L}}{\partial \boldsymbol{m}^{l}[t]} \boldsymbol{\psi}^{l}[t] \\
&+ 
\sum_{t^{\prime}=t+1}^{T}
\left(
{\color{blue}
\frac{\partial\mathcal{L}}{\partial \boldsymbol{s}^{l}[t^{\prime}]} 
\frac{\partial \boldsymbol{s}^{l}[t^{\prime}]}{\partial \boldsymbol{u}^{l}[t^{\prime }]}
}
+
\frac{\partial\mathcal{L}}{\partial \boldsymbol{m}^{l}[t^{\prime}]} \boldsymbol{\psi}^{l}[t^{\prime}] 
\right)
\prod_{t^{\prime\prime}=1}^{t^{\prime}-t} 
\left( {\color{blue}\boldsymbol{\epsilon}^{l}[t^{\prime}-t^{\prime\prime}]} +  
\frac{\partial \boldsymbol{u}^{l}[t^{\prime}-t^{\prime\prime}+1]}{\partial \boldsymbol{m}^{l}[t^{\prime}-t^{\prime\prime}]} \boldsymbol{\psi}^{l}[t^{\prime}-t^{\prime\prime}] \right) \\
=& 
{\color{blue}
\frac{\partial\mathcal{L}}{\partial \boldsymbol{s}^{l}[t]}  
\frac{\partial \boldsymbol{s}^{l}[t]}{\partial \boldsymbol{u}^{l}[t]}
}
+
\frac{\partial\mathcal{L}}{\partial \boldsymbol{m}^{l}[t]} \boldsymbol{\psi}^{l}[t] 
+ 
\sum_{t^{\prime}=t+1}^{T}
{\color{blue}
\frac{\partial\mathcal{L}}{\partial \boldsymbol{s}^{l}[t^{\prime}]} 
\frac{\partial \boldsymbol{s}^{l}[t^{\prime}]}{\partial \boldsymbol{u}^{l}[t^{\prime }]}
}
\prod_{t^{\prime\prime}=1}^{t^{\prime}-t} 
\left( {\color{blue}\boldsymbol{\epsilon}^{l}[t^{\prime}-t^{\prime\prime}]} +  
\frac{\partial \boldsymbol{u}^{l}[t^{\prime}-t^{\prime\prime}+1]}{\partial \boldsymbol{m}^{l}[t^{\prime}-t^{\prime\prime}]} \boldsymbol{\psi}^{l}[t^{\prime}-t^{\prime\prime}] \right) \\
&+
\sum_{t^{\prime}=t+1}^{T}
\frac{\partial\mathcal{L}}{\partial \boldsymbol{m}^{l}[t^{\prime}]} \boldsymbol{\psi}^{l}[t^{\prime}] 
\prod_{t^{\prime\prime}=1}^{t^{\prime}-t} 
\left( {\color{blue}\boldsymbol{\epsilon}^{l}[t^{\prime}-t^{\prime\prime}]} +  
\frac{\partial \boldsymbol{u}^{l}[t^{\prime}-t^{\prime\prime}+1]}{\partial \boldsymbol{m}^{l}[t^{\prime}-t^{\prime\prime}]} \boldsymbol{\psi}^{l}[t^{\prime}-t^{\prime\prime}] \right)
\end{aligned}
\end{equation}

Here again, {\color{blue}the gradient of the original LIF neuron is plotted in blue}. We can intuitively see that the temporal gradient contributions from the Complementary component are more significant than those from LIF. Even in the worst case all of $\prod\boldsymbol{\xi} \to 0$, $\frac{\partial\mathcal{L}}{\partial \boldsymbol{m}^{l}[t]}$ also provides the sum of all temporal gradients as shown in Figure.\eqref{aeq:dL_dm_final_expression}, like shortcut at temporal dimension. 

\newpage

\newpage
\section{The Loss Comparing between LIF and CLIF}
\label{appendix:The Loss Comparing between LIF and CLIF}

\begin{figure*}[hbt]
\centering
\subfigure[CIFAR10] {\includegraphics[width=0.19\textwidth,trim=0 0 0 0,clip]{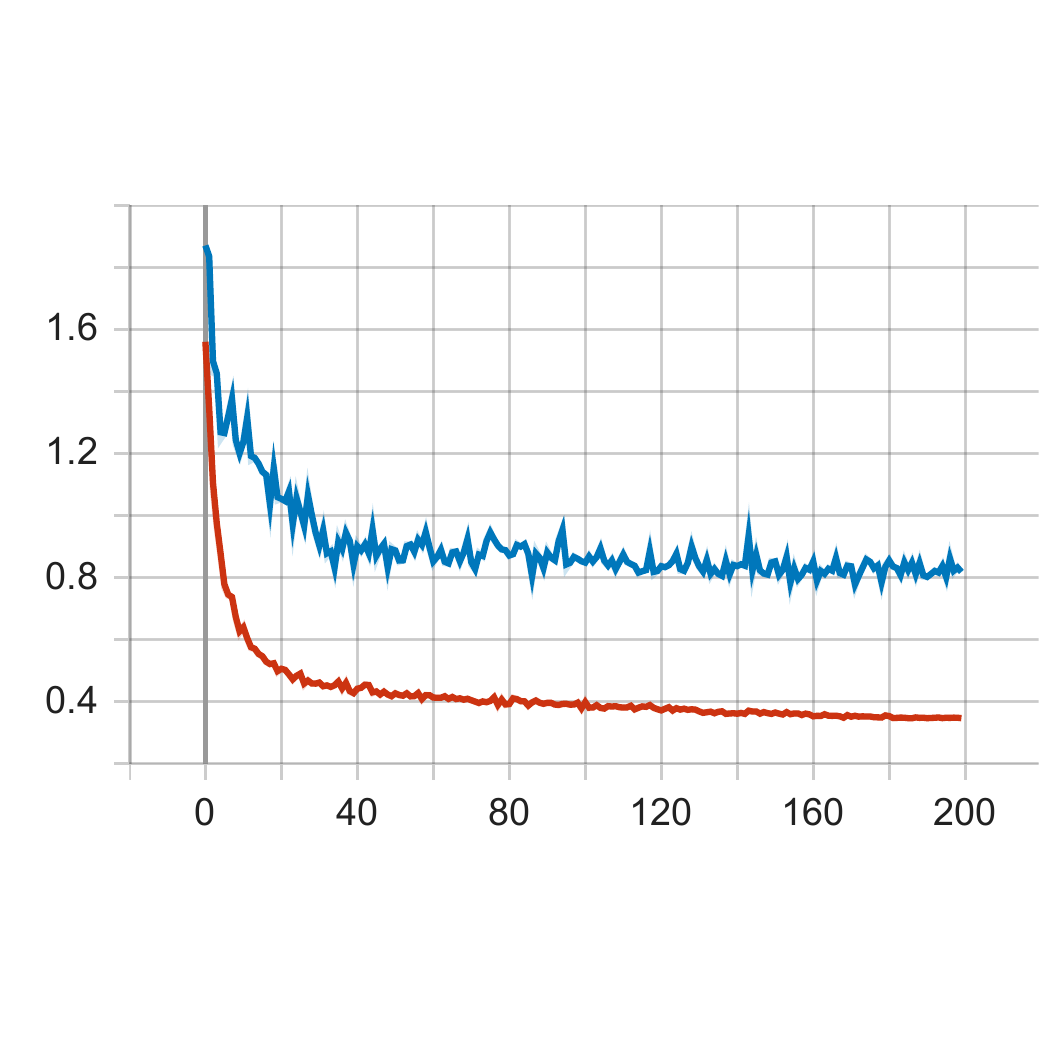}}
\subfigure[CIFAR100] {\includegraphics[width=0.19\textwidth,trim=0 0 0 0,clip]{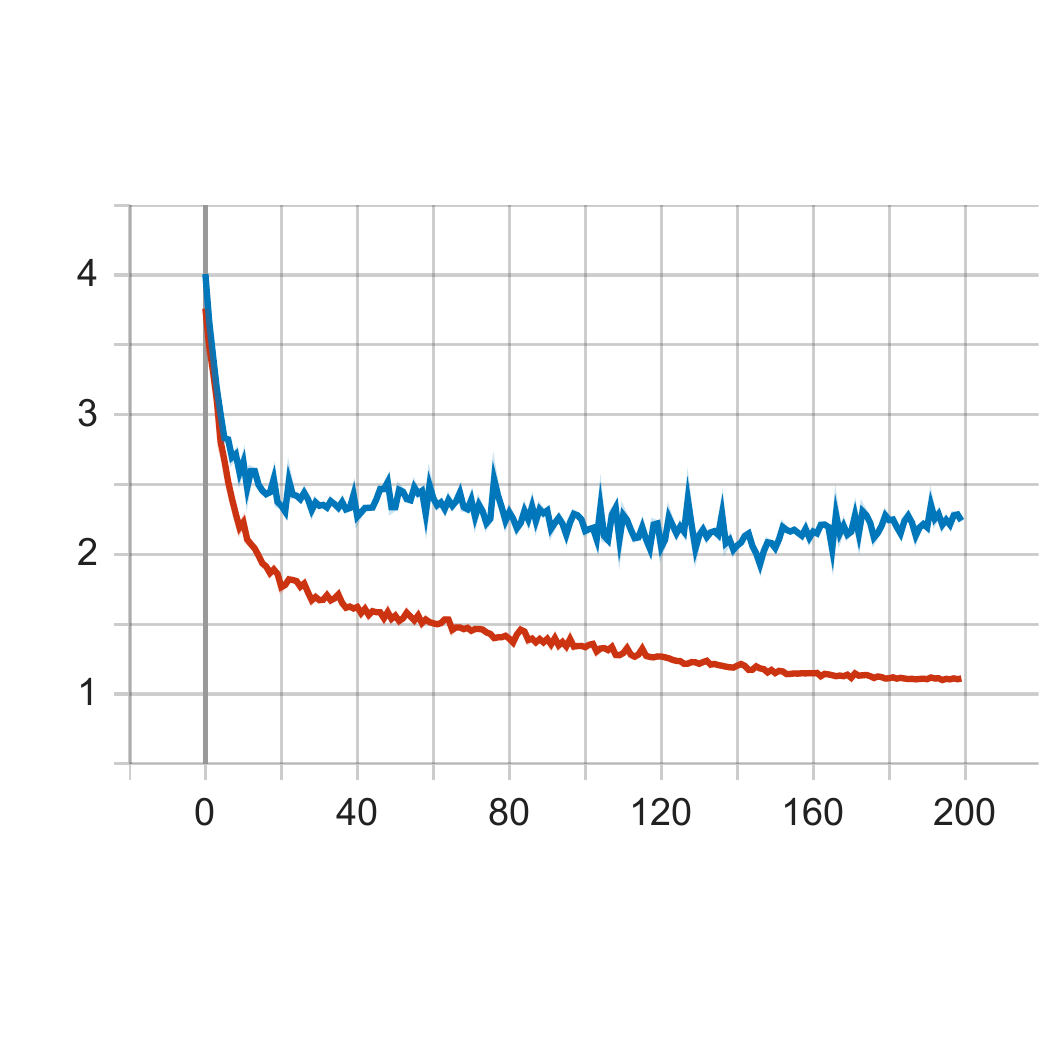}} 
\subfigure[Tiny Imagenet] {\includegraphics[width=0.19\textwidth,trim=0 0 0 0,clip]{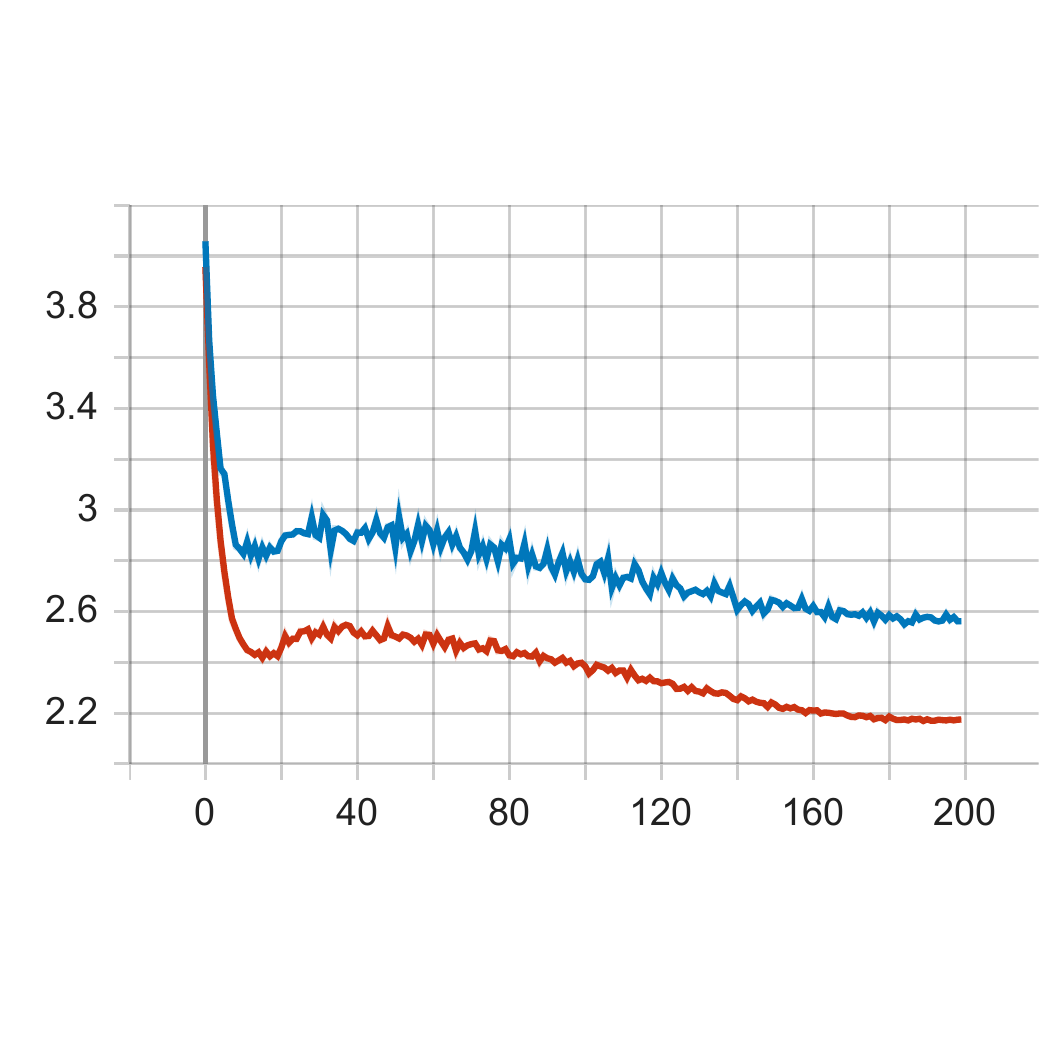}} 
\subfigure[DVSCIFAR10] {\includegraphics[width=0.19\textwidth,trim=0 0 0 0,clip]{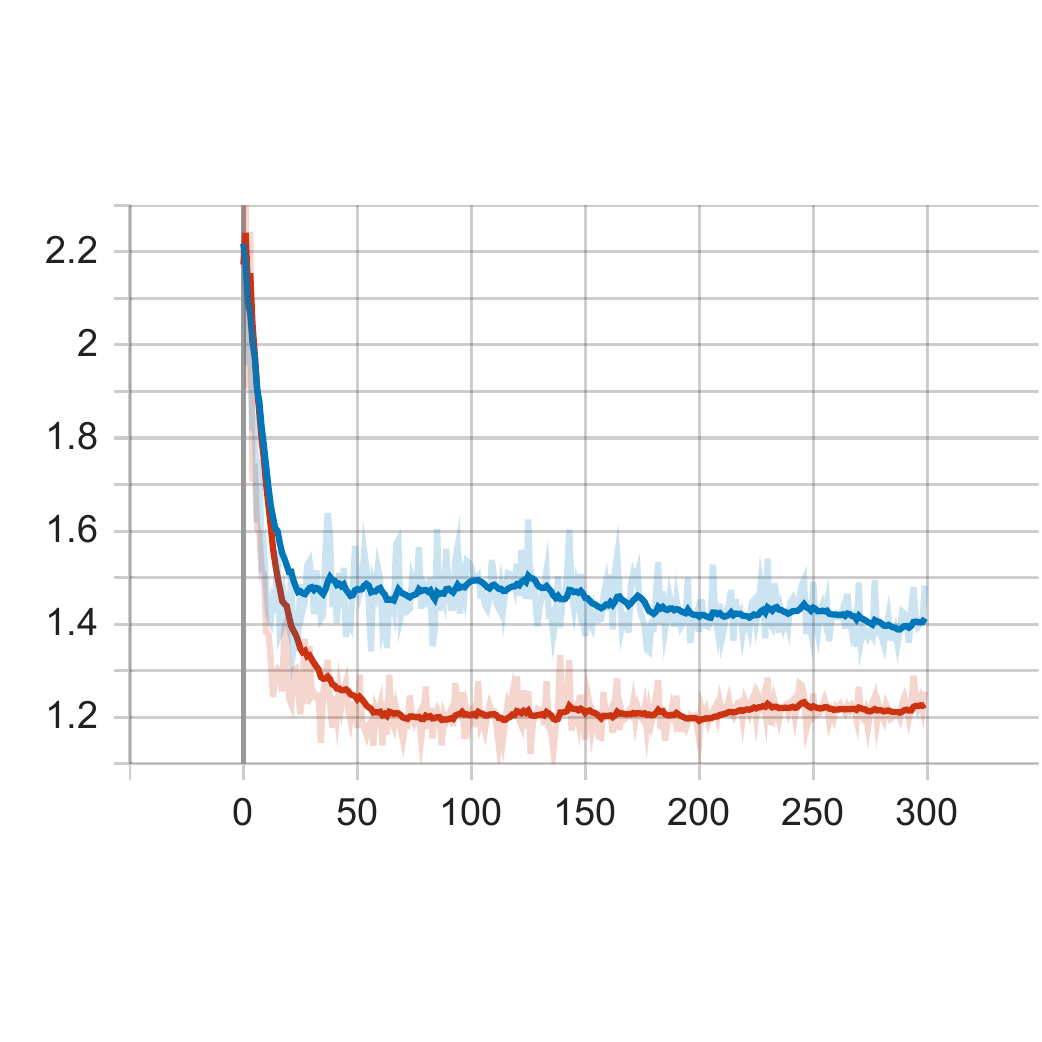}} 
\subfigure[DVSGesture] {\includegraphics[width=0.19\textwidth,trim=0 0 0 0,clip]{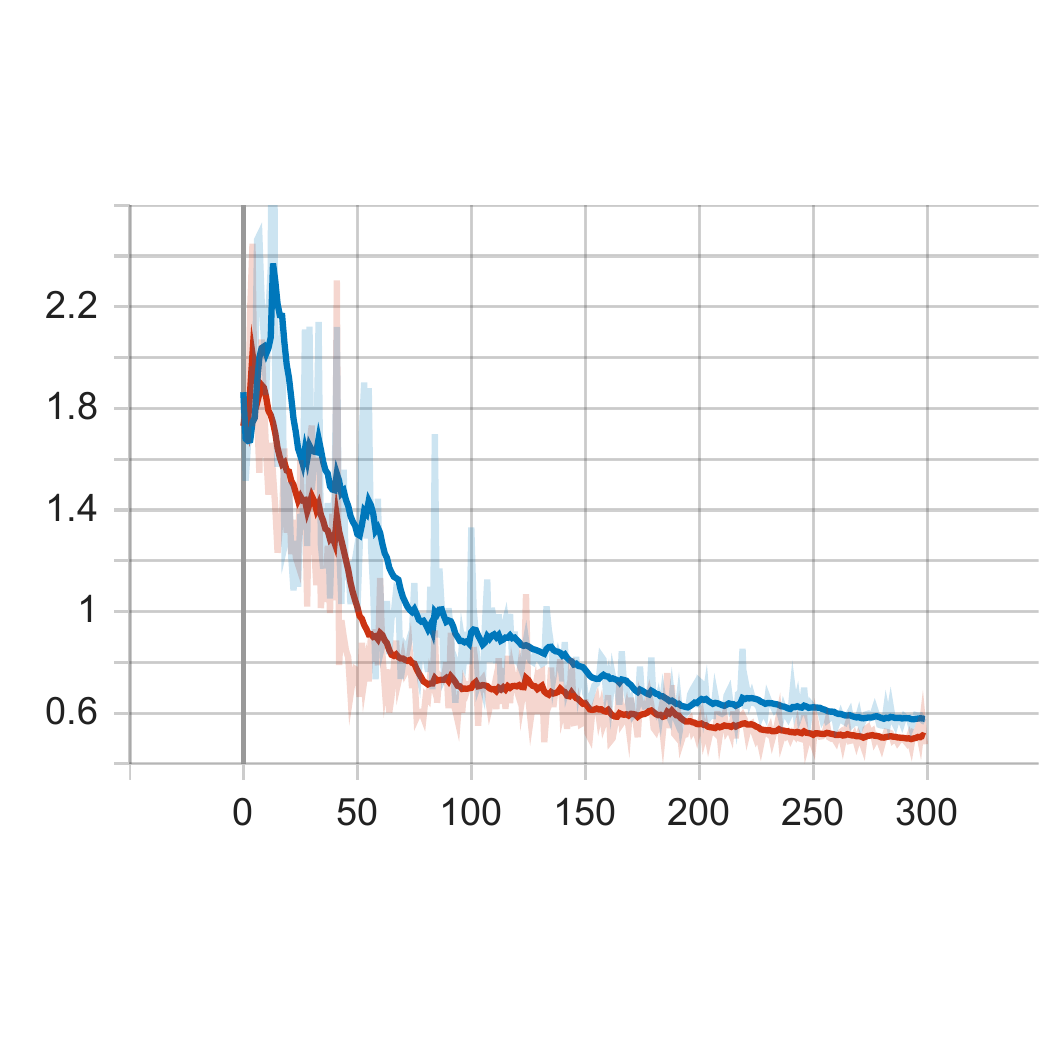}} 
\caption{Testing loss curves of the training process of \textcolor{MyBlue}{\textbf{LIF}}-based and \textcolor{MyRed}{\textbf{CLIF}}-based for each tasks.}
\label{fig:appendix:loss_compare}
\end{figure*}

Following Figure.\ref{fig:acc_vs_T} we extend the loss comparison to various tasks and network backbones. The results are shown in Figure.\ref{fig:appendix:loss_compare}, CLIF neuron’s loss converges faster than LIF’s, the converged loss is also lower. This tendency demonstrates the advantage of the CLIF neuron model.

\section{Experiment Description and Dataset Pre-processing}
\label{appendix:Dataset Description and Preprocessing}
Unless otherwise specified or for the purpose of comparative experiments, the experiments in this paper adhere to the following settings and data preprocessing: all our self-implementations use Rectangle surrogate functions with $\alpha = V_\mathrm{th} = 1$, and the decay constant $\tau$ is set to $2.0$. All random seed settings are 2022. For all loss functions, we use the TET \cite{deng2021temporal} with a 0.05 loss lambda, as implemented in \cite{meng2023towards}. The following are the detailed default setups and dataset descriptions.

\begin{table}[hb]
\centering
\caption{Training hyperparameters}
\label{tab:appendix:hyper}
\begin{tabular}{cccccc}
\toprule
Dataset       & Optimizer & Weight Dacay & Batch Size & Epoch & Learning Rate \\ \midrule
CIFAR10       & SGD       & 5e-5         & 128        & 200   & 0.1           \\ \midrule
CIFAR100      & SGD       & 5e-4         & 128        & 200   & 0.1           \\ \midrule
Tiny ImageNet & SGD       & 5e-4         & 256        & 300   & 0.1           \\ \midrule
DVSCIFAR10    & SGD       & 5e-4         & 128        & 300   & 0.05          \\ \midrule
DVSGesture    & SGD       & 5e-4         & 16         & 300   & 0.1           \\ 
\bottomrule
\end{tabular}
\end{table}

\textbf{CIFAR-10/100} The CIFAR-10 and CIFAR-100 datasets \cite{krizhevsky2009learning} contain 60,000 32×32 color images in 10 and 100 different classes respectively, with each dataset comprising 50,000 training samples and 10,000 testing samples. We normalize the image data to ensure that input images have zero mean and unit variance. For data preprocessing, we directly follow this work \cite{meng2023towards}. We apply random cropping with padding of 4 pixels on each border of the image, random horizontal flipping, and cutout \cite{devries2017improved}. Direct encoding \cite{rathi2021diet} is employed to encode the image pixels into time series, wherein pixel values are applied repeatedly to the input layer at each timestep. For the CIFAR classification task, we use Spiking-Resnet18 as the backbone.

\textbf{Tiny-ImageNet} Tiny-ImageNet contains 200 categories and 100,000 64×64 colored images for training, which is a more challenging static image dataset than CIFAR datasets. To augment Tiny-ImageNet datasets, we take the same AutoAugment \cite{cubuk2019autoaugment} as used in this work \cite{wang2023adaptive}, but we do not adopt Cutout \cite{devries2017improved}. For the Tiny-ImageNet classification task, we use Spiking-VGG13 as the backbone.

\textbf{DVSCIFAR10} The DVS-CIFAR10 dataset \cite{li2017cifar10} is a neuromorphic dataset converted from CIFAR-10 using a DVS camera. It contains 10,000 event-based images with pixel dimensions expanded to 128×128. The event-to-frame integration is handled with the SpikingJelly \cite{fang2023spikingjelly} framework. We do not apply any data augmentation for DVSCIFAR10 and the Spiking-VGG11 is used as the backbone to compare the performance.

\textbf{DVSGesture} The DVS128 Gesture dataset \cite{amir2017low} is a challenging neuromorphic dataset that records 11 gestures performed by 29 different participants under three lighting conditions. The dataset comprises 1,342 samples with an average duration of 6.5 ± 1.7 s and all samples are split into a training set (1208 samples) and a test set (134 samples). We follow the method described in \cite{fang2021incorporating} to integrate the events into frames. The event-to-frame integration is handled with the SpikingJelly \cite{fang2023spikingjelly} framework. We do not applied any data augmentation for DVSGesture and the Spiking-VGG11 is used as the backbone to compare the performance.

\section{Evaluation of Fire Rate and Energy Consumption}
\label{appendix:The Comparing of Computing Efficiency}

\begin{figure*}[hbt]
\centering
\subfigure[CIFAR10] {\includegraphics[width=0.19\textwidth,trim=0 0 0 0,clip]{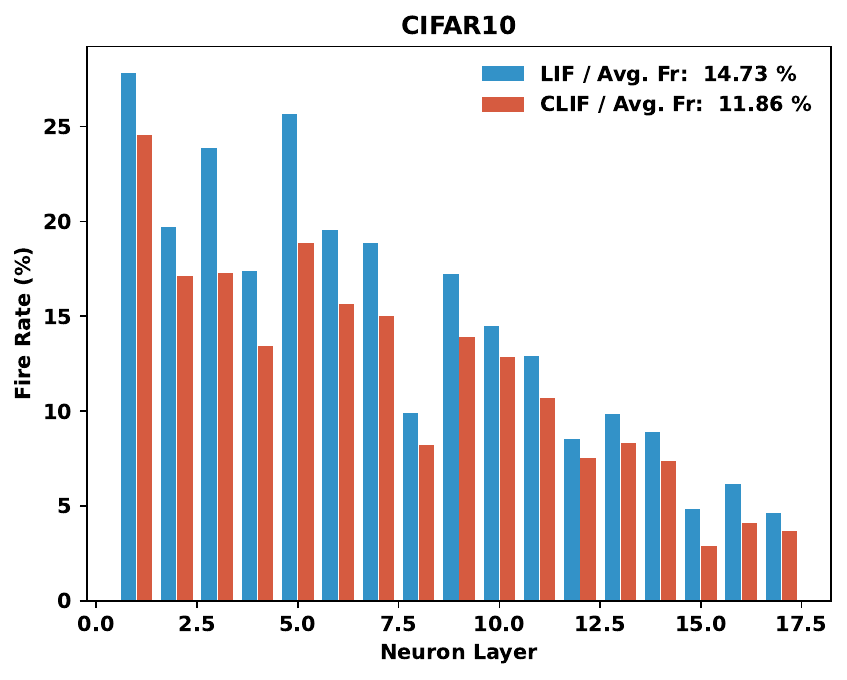}}
\subfigure[CIFAR100] {\includegraphics[width=0.19\textwidth,trim=0 0 0 0,clip]{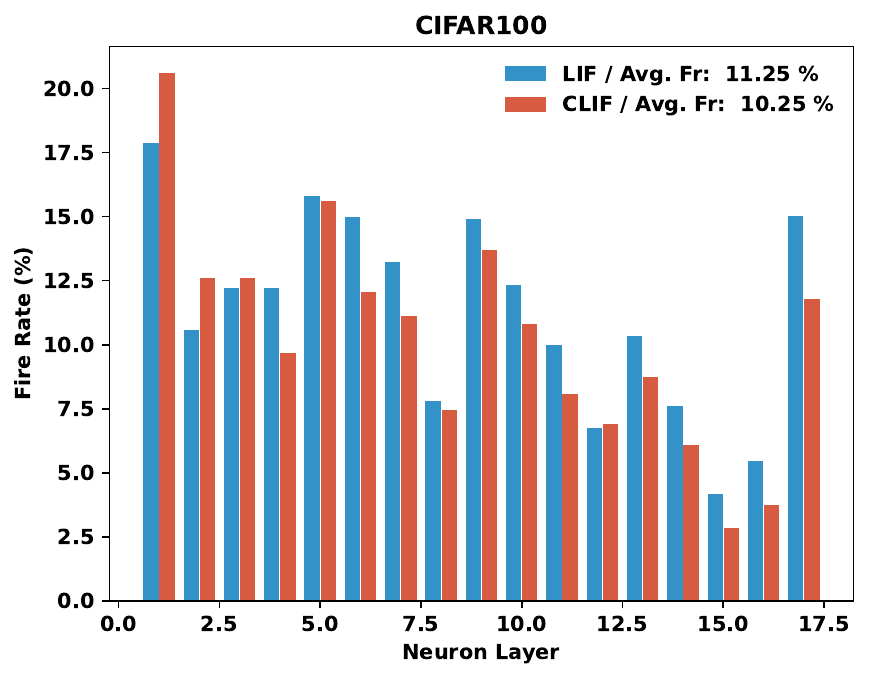}} 
\subfigure[Tiny Imagenet] {\includegraphics[width=0.19\textwidth,trim=0 0 0 0,clip]{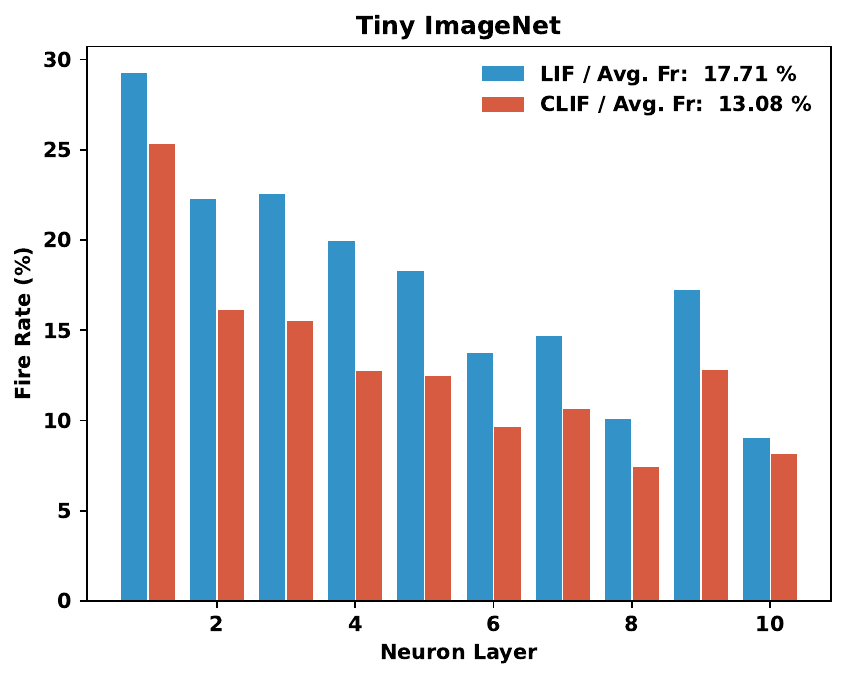}} 
\subfigure[DVSCIFAR10] {\includegraphics[width=0.19\textwidth,trim=0 0 0 0,clip]{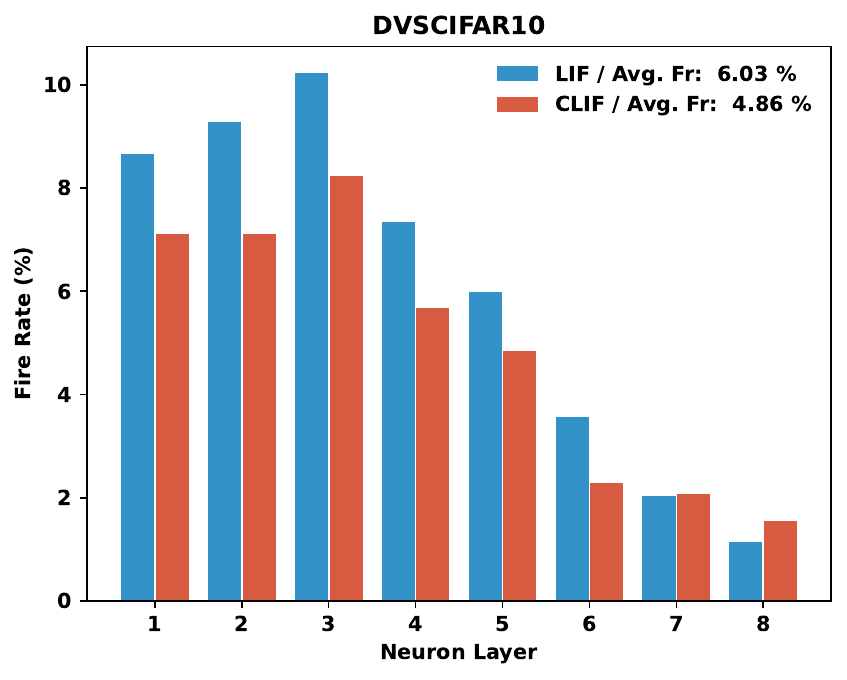}} 
\subfigure[DVSGesture] {\includegraphics[width=0.19\textwidth,trim=0 0 0 0,clip]{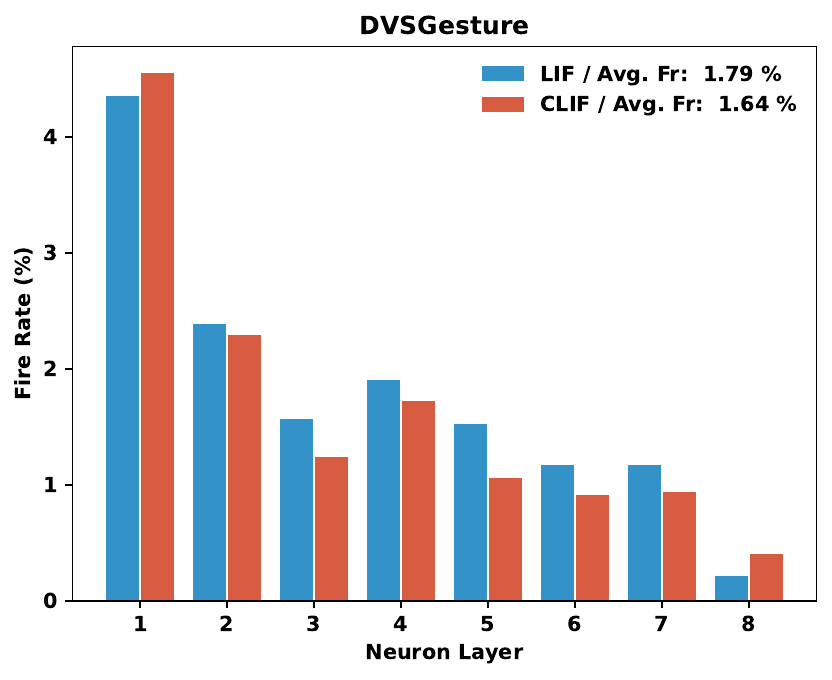}} 
\caption{The fire rate statistics after training of \textcolor{MyBlue}{\textbf{LIF}}-based and \textcolor{MyRed}{\textbf{CLIF}}-based for each tasks.}
\label{fig:appendix:fire_rate}
\end{figure*}

We calculate the fire rate as well as the energy efficiency of all models for five tasks. As shown in Figure.\ref{fig:appendix:fire_rate}, the average fire rate of the CLIF model is lower than that of the LIF model. This lower fire rate results in fewer synaptic operations, as evidenced in Table.\ref{tab:appendix:energy effiency}. 

For the evaluation of energy consumption, we follow the convention of the neuromorphic computing community by counting the total synaptic operations (SOP) to estimate the computation overhead of SNN models and compare it to the energy consumption of the ANN counterpart, as done in \cite{zhou2022spikformer, yao2024spike}. Specifically, the SOP with MAC presented in ANNs is constant given a specified structure. However, the SOP in SNN varies with spike sparsity. For SNNs, since the input is binary, the synaptic operation is mostly accumulation (ACs) instead of multiply and accumulation (MACs). ACs is defined as
\begin{equation}
AC_s = \sum_{t=1}^{T}\sum_{l=1}^{L-1}\sum_{i=1}^{N^l} f_i^l s_i^l[t]
\end{equation}
where fan-out $f_i^l$ is the number of outgoing connections to the subsequent layer, and $N^l$ is the neuron number of the $l$-th layer. For ANNs, the similar synaptic operation MACs with more expensive multiply-accumulate is defined as:
\begin{equation}
MAC_s = \sum_{l=1}^{L-1}\sum_{i=1}^{N^l} f_i^l
\end{equation}
Here, we select all the testing datasets and estimate the average SOP for SNNs. Meanwhile, we measure 32-bit floating-point ACs by 0.9 pJ per operation and 32-bit floating-point MAC by 4.6 pJ per operation, as done in \cite{han2015learning}. All the results are summarized in the Table.\ref{tab:appendix:energy effiency}, SNN has a significant energy consumption advantage over ANNs. Notably, the ACs operation of CLIF are considerably less than those of LIF, attributable to the lower fire rate. In contrast, the MAC operations of CLIF exceed those of LIF due to the increased number of floating-point operations, a result of the Complementary component introduced in CLIF. The final results indicate that CLIF achieves comparable performance to ANNs models while maintaining similar total energy efficiency to LIF.

\begin{table*}[h]
\centering
\caption{The energy consumption of synaptic operation for different tasks with the whole testing datasets.}
\label{tab:appendix:energy effiency}
\begin{tabular}{cccccccc} %
\toprule
 &
  \textbf{Neuron} &
  \textbf{T} &
  \textbf{Parameters(M)} &
  \textbf{ACs (M)} &
  \textbf{MACs (M)} &
  \textbf{SOP Energy (µJ)} & \\
\midrule
 & \text{ReLU} & 1  & 11.2 & 0     & 557.65 & 2565.19  \\
 & \text{CLIF} & 6  & 11.2 & 68.66 & 5.12   & 85.346  \\
\multirow{-3}{*}{\makecell{\textbf{CIFAR10} \\ (ResNet18)}}  & \text{LIF}  & 6  & 11.2 & 84.86 & 2.89   & 89.668   \\
\midrule
 & \text{ReLU} & 1  & 11.2 & 0     & 557.7  & 2565.42   \\
 & \text{CLIF} & 6  & 11.2 & 55.58 & 5.16   & 73.758   \\
\multirow{-3}{*}{\makecell{\textbf{CIFAR100} \\ (ResNet18)}} & \text{LIF}  & 6  & 11.2 & 60.28 & 2.93   & 67.73   \\
\midrule
  & \text{ReLU} & 1  & 14.4 & 0     & 922.56 & 4243.776  \\
  & \text{CLIF} & 6  & 14.4 & 102.1 & 282.83 & 1392.908  \\
\multirow{-3}{*}{\makecell{\textbf{Tiny ImageNet} \\ (VGG13)}} &
  \text{LIF} &
  6 &
  14.4 &
  135.25 &
  278.84 &
  1404.389  \\
\midrule
   & \text{CLIF} & 20 & 9.5  & 19.16 & 1090   & 5031.244  \\
\multirow{-2}{*}{\makecell{\textbf{DVSGesture} \\ (VGG11)}}  & \text{LIF}  & 20 & 9.5  & 25.09 & 1080   & 4990.581 \\
\midrule
      & \text{CLIF} & 10 & 9.5  & 12.02 & 153.87 & 718.62   \\
\multirow{-2}{*}{\makecell{\textbf{DVSCIFAR10} \\ (VGG11)}}  & \text{LIF}  & 10 & 9.5  & 14.65 & 152.5  & 714.685  \\
\bottomrule
\end{tabular}%
\end{table*}

\newpage 

{\color{black}
To comprehensively and fairly evaluate the energy consumption \cite{dampfhoffer2022snns}, we recalculated and analyzed the energy consumption of the proposed CLIF neuron in more detail in Table \ref{tab:appendix:total energy effiency}. We considered memory read and write operations, as well as the data addressing process, as done in \cite{lemaire2022analytical}. As shown in Table \ref{tab:appendix:total energy effiency}, the memory accesses are actually the dominant factor in energy consumption for SNN. Although the hidden states of LIF and CLIF contribute significantly to the read and write energy consumption of the membrane potential, the sparsity of spikes also greatly reduces the parameters and synaptic operations. Therefore, the energy consumption of LIF and CLIF is still much lower than that of ANN. The detailed computing process can be found in the open-source code.
}

\begin{table}[ht!]
\centering
\caption{The total energy consumption for different tasks. The neuron and time step are the same as those in Table 6.}
\label{tab:appendix:total energy effiency}
\begin{tabular}{ccccccc}
\toprule
 &
  \multicolumn{3}{c}{\textbf{Mem. Read \& Write}} &
  \multirow{2}{*}{\textbf{\begin{tabular}[c]{@{}c@{}}Synaptic \&\\  Neuron Op. (mJ)\end{tabular}}} &
  \multirow{2}{*}{\textbf{Addr. (µJ)}} &
  \multirow{2}{*}{\textbf{Total (mJ)}} \\
                                       & \begin{tabular}[c]{@{}c@{}}Membrane \\ Potential (mJ)\end{tabular}
                                       & Parameters (mJ) & In / Out (mJ) &        &         &                   \\
\midrule
\multirow{3}{*}{\textbf{CIFAR10}}      & 0                  & 54.9688         & 54.9357     & 1.7573 & 0.1145  & \textbf{111.6619} \\
                                       & 22.9987                 & 11.4994         & 0.0013      & 0.0190 & 12.1394 & \textbf{34.5304}  \\
                                       & 55.5172                 & 9.2529          & 0.0007      & 0.0389 & 19.5184 & \textbf{64.8293}  \\
\midrule
\multirow{3}{*}{\textbf{CIFAR100}}     & 0                  & 54.9735         & 54.9357     & 1.7574 & 0.1145  & \textbf{111.6667} \\
                                       & 16.8666                 & 8.4337          & 0.0004      & 0.0171 & 8.8427  & \textbf{25.3267}  \\
                                       & 46.8265                 & 7.8048          & 0.0004      & 0.0380 & 16.4053 & \textbf{54.6861}  \\
\midrule
\multirow{3}{*}{\textbf{TinyImagenet}} & 0                  & 91.9065         & 91.3834     & 2.9379 & 0.1686  & \textbf{186.2279} \\
                                       & 37.7700                 & 18.9076         & 0.0050      & 1.1754 & 19.8379 & \textbf{57.8778}  \\
                                       & 84.8900                 & 14.1756         & 0.0028      & 1.0986 & 29.7983 & \textbf{100.1968} \\
\midrule
\multirow{2}{*}{\textbf{DVSCIFAR10}}   & 5.7398                  & 2.8701          & 0.0003      & 0.0212 & 2.9340  & \textbf{8.6343}   \\
                                       & 14.0891                 & 2.3484          & 0.0002      & 0.0451 & 4.7976  & \textbf{16.4876}  \\
\midrule
\multirow{2}{*}{\textbf{DVSGesture}}   & 13.9212                 & 6.9606          & 0.0003      & 0.1682 & 7.0728  & \textbf{21.0574}  \\
                                       & 38.3082                 & 6.3848          & 0.0003      & 0.4794 & 12.9767 & \textbf{45.1856} \\
\bottomrule
\end{tabular}
\end{table}

\newpage

\newpage

{\color{black}
In addition, it is feasible to train a model using CLIF and subsequently deploy it or inference with LIF. We take pre-trained models of CLIF and LIF (Resnet18 with T=6) to perform inference on the CIFAR10 and CIFAR100 tasks. To compensate for CLIF’s enhanced reset process, we employ a hard reset with a bias as a hyperparameter. As can be seen in Table, this approach leads to an inference accuracy that surpasses that of a model directly trained with LIF.
}

\begin{table}[ht]
\caption{Directly convert the pre-trained CLIF/LIF model to an LIF neuron for inference.}
\begin{tabular}{l|ccccccc}
\toprule
\textbf{CIFAR10} &
  \textbf{Soft Reset} &
  \textbf{Hard Reset} &
   &
   &
   &
   &
  \textbf{} \\
\midrule
\textbf{Reset Value} &
  None &
  0 &
  -0.02 &
  -0.04 &
  -0.06 &
  -0.08 &
  -0.1 \\
\midrule
{\textbf{CLIF pretrained (95.41\%)}} &
  92.95 \% &
  93.41 \% &
  94.18 \% &
  94.54 \% &
  \textbf{95.08 \%} &
  94.84 \% &
  94.72 \% \\
\midrule
{\textbf{LIF pretrained (94.51\%)}} &
  94.51 \% &
  84.05 \% &
  76.68 \% &
  66.08 \% &
  52.16 \% &
  38.00 \% &
  27.04 \% \\
\toprule
\toprule
\textbf{CIFAR100} &
  \textbf{Soft Reset} &
  \textbf{Hard Reset} &
  \textbf{} &
  \textbf{} &
  \textbf{} &
   &
  \textbf{} \\
\midrule
\textbf{Reset Value} &
  None &
  0 &
  -0.02 &
  -0.04 &
  -0.06 &
  -0.08 &
  -0.1 \\
\midrule
{\textbf{CLIF pretrained (78.36 \%)}} &
  68.72 \% &
  73.04 \% &
  74.64 \% &
  76.63 \% &
  76.55 \% &
  \textbf{77.00 \%} &
  76.54 \% \\
\midrule
{\textbf{LIF pretrained (76.23 \%)}} &
  76.23 \% &
  47.74 \% &
  37.04 \% &
  27.56 \% &
  19.83 \% &
  13.22 \% &
  8.77 \% \\
\bottomrule
\end{tabular}
\end{table}


\end{document}